\documentclass[11pt]{article}

\usepackage[preprint]{acl}

\usepackage{times}
\usepackage{latexsym}

\usepackage[T1]{fontenc}
\usepackage[utf8]{inputenc}

\usepackage{microtype}
\usepackage{inconsolata}

\usepackage{graphicx}
\usepackage{caption}
\usepackage{subcaption}
\usepackage{booktabs}
\usepackage{multirow}
\usepackage{makecell}
\usepackage{tabularx}
\usepackage{array}
\usepackage{colortbl}
\usepackage{adjustbox}
\usepackage{threeparttable}

\usepackage{amsmath,amssymb,amsfonts}
\usepackage{mathtools}

\usepackage{algorithm}
\usepackage{algpseudocode}

\usepackage{enumitem}

\usepackage{tikz}
\usetikzlibrary{positioning,arrows.meta,calc,shapes.geometric,fit,backgrounds,decorations.pathreplacing,patterns}
\usepackage{pgfplots}
\pgfplotsset{compat=1.18}
\usepgfplotslibrary{polar}

\usepackage{xcolor}
\usepackage[capitalize,nameinlink]{cleveref}

\usepackage{xspace}

\newcommand{\method}{\textit{OmniRetriever}}

\newcommand{\bench}{\textit{OmniRetriever-Bench}}
\newcommand{\model}{\textit{OmniRetriever-7B}}
\providecommand{\eg}{\textit{e.g.}}

\providecommand{\cmark}{$\checkmark$}

\title{\method{}: Any-to-Any Audio-Video-Text Retrieval\\
       via Fusion-as-Teacher Distillation}

\author{
  Yunze Liu \quad Chi-Hao Wu \quad Enmin Zhou \quad Junxiao Shen \\
  Memories.ai Research \\
  Project Page: \href{https://yunzeliu.github.io/OmniRetriever/}{OmniRetriever}
}

\begin{document}
\maketitle

\begin{figure*}[!t]
\centering
\includegraphics[width=\linewidth]{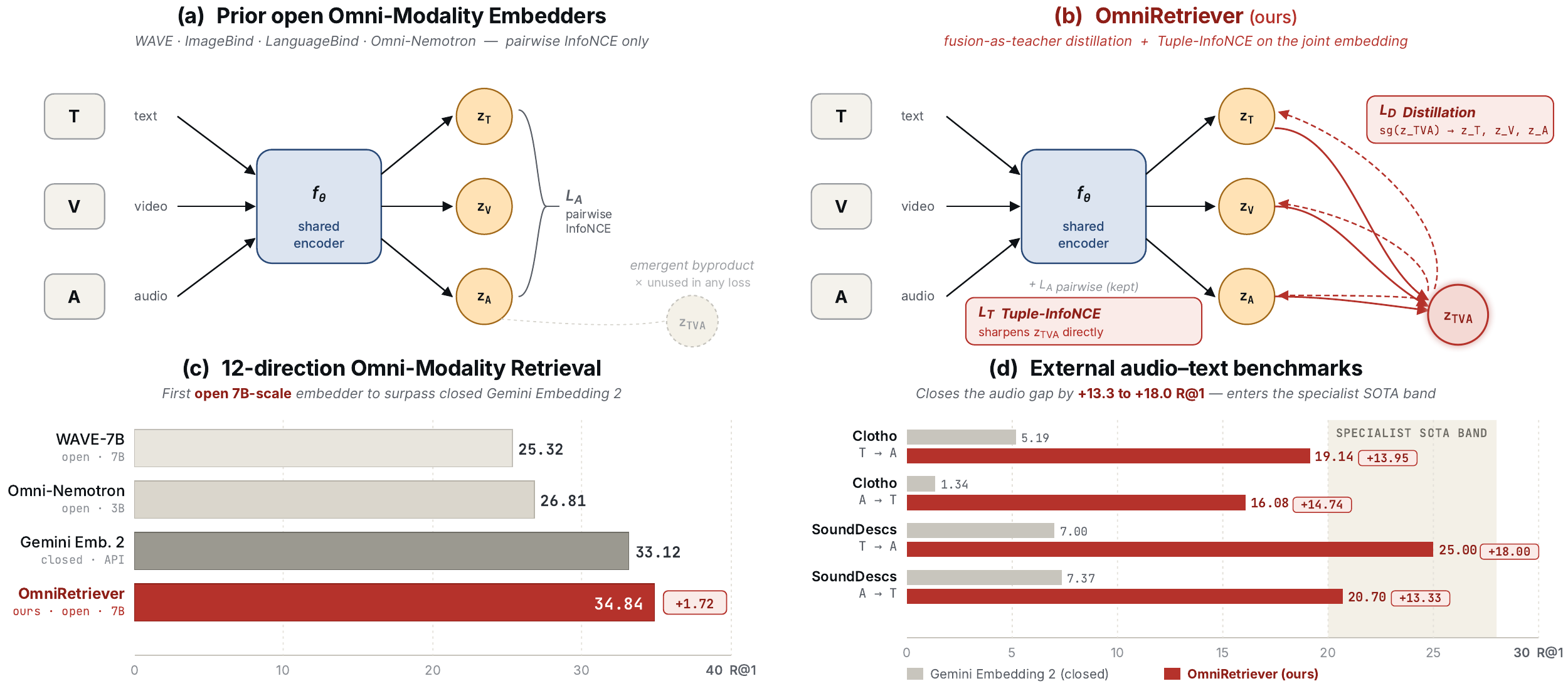}
\caption{\textbf{Method overview.}
\method{} uses the joint embedding $\mathbf{z}_{TVA}$, which is
unused by pairwise training (a), as a supervision target (b)
via fusion-as-teacher distillation $\mathcal{L}_D$ and a
Tuple-InfoNCE term $\mathcal{L}_T$. This yields a new open
result on 12-direction AVT retrieval (c) and a $13.3$ to
$18.0$ R@1 gain over Gemini Embedding~2 on external
audio--text benchmarks (d).}
\label{fig:teaser}
\end{figure*}

\begin{abstract}
Unified multimodal embedding spaces have become the standard
interface for cross-modal retrieval and multimodal RAG, and recent audio-video-text (AVT) encoders extend this setting to three modalities. Such encoders can produce joint $(T,V,A)$
embedding whenever all three modalities are available, but standard pairwise InfoNCE objectives leave this signal unused
during training. We close this
gap with \textbf{fusion-as-teacher distillation}, which
treats a stop-gradient copy of the fused embedding as a
teacher signal for the single-modal embedding, paired
with a \textbf{Tuple-InfoNCE} term that supervises the
fused embedding directly. We instantiate this objective as
\textbf{\model{}}. Across six zero-shot retrieval benchmarks,
\model{} surpasses the closed-source Gemini Embedding~2 by
$13.3$--$18.0$ R@1 on Clotho and SoundDescs, and reaches
the contemporary zero-shot specialist band of open
video--text encoders on MSR-VTT and MSVD. To stress-test
joint representations, we further release
\textbf{\bench{}}, a 12-direction AVT retrieval benchmark
totaling $3{,}782$ triples; on it \model{} attains AVG-all
$34.84$, improving over Gemini Embedding~2 by $+1.72$ and
over the best prior open-source AVT method by $+8.03$. Model weights, datasets, and code will be
released.
\end{abstract}

\section{Introduction}\label{sec:intro}

\noindent Cross-modal retrieval relies on encoders that map
queryable modalities into a single shared embedding space. The
canonical CLIP recipe~\cite{radford2021learning}
trains two separate towers and aligns them with pairwise
InfoNCE. Recent omni-modal systems extend this to three
modalities of $\{T,V,A\}$ and split into two architectural
families. \emph{Multi-encoder} systems such as
ImageBind~\cite{girdhar2023imagebind} and
LanguageBind~\cite{zhu2024languagebind} keep separate
per-modality encoders and align them pairwise to a fixed image
or language anchor. \emph{Unified} encoders such as
WAVE~\cite{tang2026wave} and
Omni-Embed-Nemotron~\cite{nvidia2025nemotron} instead route
all three modalities through one backbone and produce a joint
embedding $\mathbf{z}_{TVA}$ on a single joint forward pass.

Unified encoders are particularly attractive for AVT retrieval
because $\mathbf{z}_{TVA}$ is exactly the representation that
a dual-modal or full-tuple query (\eg, $T{+}V \to A$ or
$A{+}T{+}V \to T$) needs at inference. Yet the standard
training recipe for unified AVT encoders never invokes this
joint forward as a supervision signal: WAVE and
Omni-Embed-Nemotron both optimise three pairwise InfoNCE
losses on the single-modal sub-encoders only, and the joint
output is computed only at inference time.

The result is that the single-modal sub-encoders of current
unified AVT systems are trained in isolation, with no signal
about their cross-modal neighbours. Empirically, this gap is
sharpest on audio-anchored retrieval, where the modality-to-text
co-occurrence is the weakest in standard training data. On Clotho A$\to$T,
closed Gemini Embedding~2 reaches R@1 $=1.34$ and the best
open omni-modal system, Omni-Embed-Nemotron, reaches $3.5$,
while CLAP-family audio--text
specialists~\cite{wu2023laionclap,mei2023wavcaps,niizumi2025m2dclap}
trained on $(T,A)$ pairs alone reach $25$ to $26$ on the same
direction. The same mechanism limits all twelve any-to-any
directions over $\{T,V,A\}$ (six single-modal, six dual-modal)
that a practical AVT retriever has to serve.

We close this gap by using the joint embedding itself as the
supervision signal. \textbf{Fusion-as-teacher distillation}
$\mathcal{L}_D$ takes a stop-gradient copy of $\mathbf{z}_{TVA}$
as the teacher for the single-modal sub-encoders: each of
$\mathbf{z}_T$, $\mathbf{z}_V$, $\mathbf{z}_A$ is pulled toward
$\mathbf{z}_{TVA}$ by InfoNCE. Because the teacher is the same
backbone consumed jointly rather than an external encoder, the
sub-encoders receive the cross-modal context that no unimodal
teacher provides, at the cost of one additional joint forward
pass per step. A complementary Tuple-InfoNCE
refinement~\cite{liu2021contrastive,liu2020p4contrast}
$\mathcal{L}_T$ supervises $\mathbf{z}_{TVA}$ itself with
modality-cycled hard negatives, preventing the joint vector
from collapsing onto the strongest pair gradient (in practice
$T$--$V$).

We instantiate this recipe as \model{}, an open 7\,B AVT
retriever. On six standard zero-shot retrieval benchmarks,
\model{} improves over closed Gemini Embedding~2 by $13$ to
$18$ R@1 on all four audio--text directions of Clotho and
SoundDescs, reaches the zero-shot CLAP-family specialist band
on Clotho T$\to$A within $\sim$$2$ R@1 of SOTA, and matches
the contemporary open zero-shot specialist band on MSR-VTT
and MSVD. To probe the six dual-modal directions
($T\!\leftrightarrow\!AV$, $A\!\leftrightarrow\!TV$,
$V\!\leftrightarrow\!AT$) that no public retrieval benchmark
currently evaluates, we additionally release
\textbf{\bench{}}, a 12-direction AVT retrieval pool of
$3{,}782$ held-out triples. On \bench{}, \model{} reaches
AVG-all $34.84$, $+1.72$ over Gemini Embedding~2 and $+8.03$
over Omni-Embed-Nemotron. A cross-backbone replication
(\Cref{sec:supp:xbackbone}) reproduces the dominant
$\mathcal{L}_D$ contribution at smaller scale, indicating that
the recipe is not tied to a particular backbone.

\begin{figure*}[t]
\centering
\includegraphics[width=\linewidth]{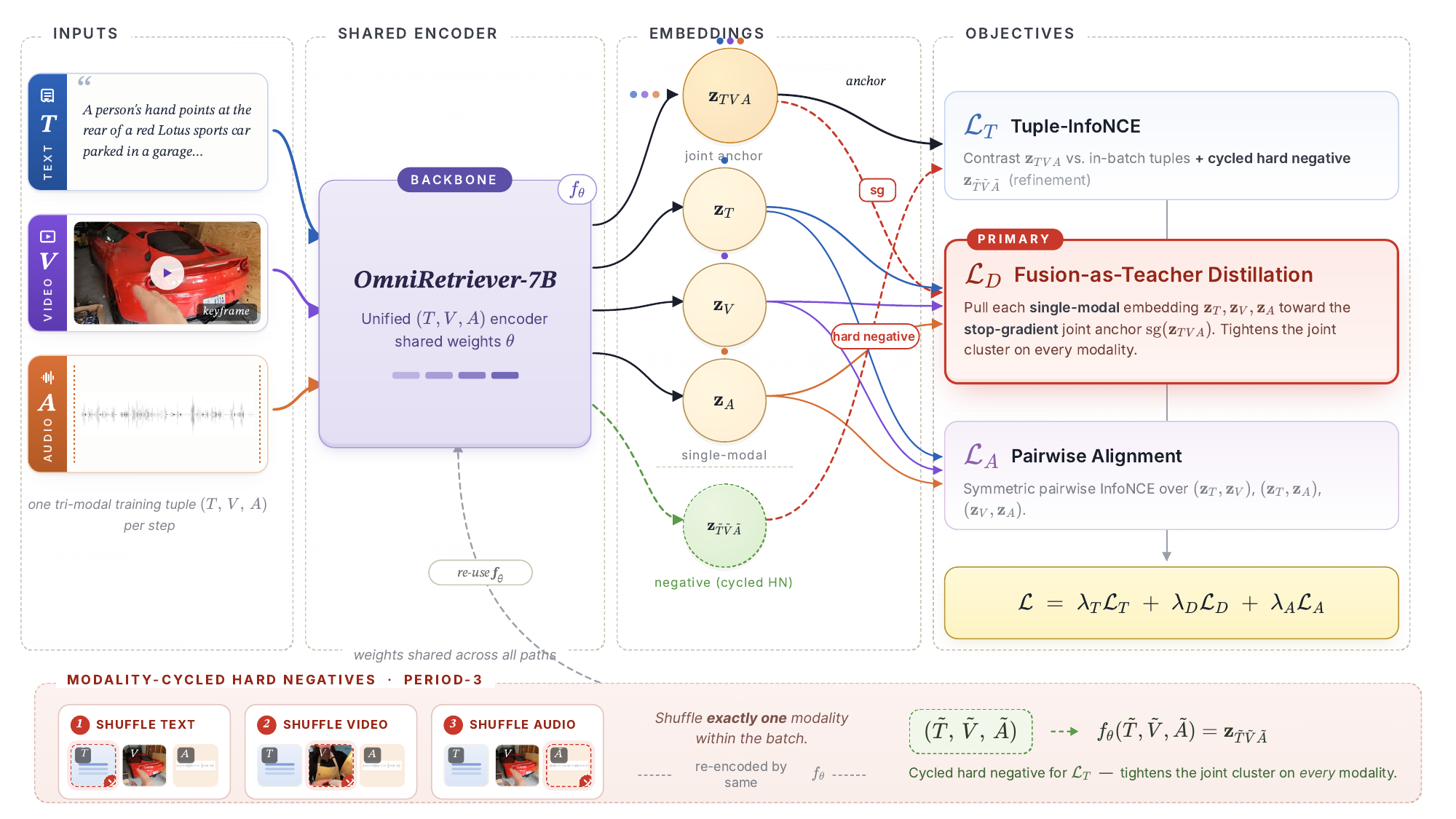}
\caption{\textbf{\method{} training overview.}
A shared encoder $f_\theta$ consumes the three modalities
jointly, producing the full-modal anchor $\mathbf{z}_{TVA}$, or
individually, producing $\mathbf{z}_T,\mathbf{z}_V,\mathbf{z}_A$.
$\mathcal{L}_D$ (fusion-as-teacher distillation, primary;
\Cref{sec:fat}) pulls each single-modality embedding toward a
stop-gradient copy of $\mathbf{z}_{TVA}$.
$\mathcal{L}_T$ (Tuple-InfoNCE refinement; \Cref{sec:tuple})
supervises $\mathbf{z}_{TVA}$ against the in-batch tuple grid
plus a modality-cycled hard negative
$\mathbf{z}_{\tilde{T}\tilde{V}\tilde{A}}$
(\Cref{eq:tuple-nce}).
$\mathcal{L}_A$ (pairwise alignment; \Cref{sec:pairwise}) ties
pairs of single-modality embeddings via symmetric InfoNCE. At
each step the hard negative perturbs one of $T,V,A$ on a
period-$3$ schedule.}
\label{fig:method}
\end{figure*}

\textbf{Contributions.}
\begin{itemize}[leftmargin=*,nosep,topsep=2pt]
\item Fusion-as-teacher distillation: the joint multimodal
embedding $\mathbf{z}_{TVA}$ of a unified AVT encoder is used
to supervise its own single-modal sub-encoders.
$\mathcal{L}_D$ alone gives the dominant single-loss gain;
the Tuple-InfoNCE term $\mathcal{L}_T$ further improves the
$A\!\leftrightarrow\!V$ routes.
\item \model{}, an open 7\,B AVT retriever. It improves over
closed Gemini Embedding~2 by $13$ to $18$ R@1 on Clotho and
SoundDescs and reaches the zero-shot audio--text specialist
band on Clotho T$\to$A within $\sim$$2$ R@1 of SOTA; on the
video--text side it matches the contemporary open zero-shot
specialist band on MSR-VTT and MSVD.
\item \bench{}, a 12-direction AVT retrieval pool of
$3{,}782$ held-out triples covering all six single- and six
dual-modal directions, the first public benchmark to evaluate
dual-modal AVT queries.
\end{itemize}

\section{Related Work}\label{sec:related}

\noindent\textbf{Pairwise contrastive vision–language alignment.}
CLIP~\cite{radford2021learning} and
ALIGN~\cite{jia2021scaling} establish image–text InfoNCE as
the standard recipe. Subsequent work scales the data
pipeline~\cite{schuhmann2022laion5b,fang2024dfn,xu2024metaclip},
replaces softmax with the sigmoid
loss~\cite{zhai2023siglip,tschannen2025siglip2}, and re-anchors
with vision foundation
models~\cite{dinov3,vjepa2,aimv2}. A parallel line of work
turns MLLMs into
encoders~\cite{wang2024e5mistral,behnamghader2024llm2vec,lee2024nvembed}.
Multimodal extensions
(VLM2Vec~\cite{jiang2025vlm2vec}, GME~\cite{zhang2024gme},
MM-Embed~\cite{lin2025mmembed},
Qwen3-VL-Embedding~\cite{alibaba2026qwen3vl}) fine-tune
MLLMs with contrastive supervision over LLM-synthesized
triplets. All apply pairwise InfoNCE per modality combination.
Our \textsc{Pairwise} baseline follows this recipe.

\noindent\textbf{Omni-modal embedding.}
ImageBind~\cite{girdhar2023imagebind} and
LanguageBind~\cite{zhu2024languagebind} align $N\!\geq\!3$
modalities pairwise to a fixed image or language anchor.
AudioCLIP~\cite{guzhov2022audioclip} and
VATT~\cite{akbari2021vatt} optimize three pairwise InfoNCE
losses over video, audio, and text.
VAST~\cite{chen2023vast} adds a 4-way alignment objective on
video–audio–subtitle–caption quadruples.
WAVE~\cite{tang2026wave} extends Qwen2.5-Omni with hierarchical
visual fusion and a dual audio encoder but still adds
two-modality InfoNCE losses, as does
Omni-Embed-Nemotron~\cite{nvidia2025nemotron}. In all of these
systems, the joint multimodal embedding is produced as a side
effect of pairwise training and never used as a supervision
target inside the same backbone.



\section{Method}\label{sec:method}

Let $\mathcal{D}=\{x_i\}$ be an AVT dataset where each
$x_i\!=\!(x_i^{(T)},x_i^{(V)},x_i^{(A)})$ carries text, video, and
audio. A unified embedder $f_\theta : \mathcal{X} \to \mathbb{R}^d$
produces single-modal embeddings
$\mathbf{z}_i^{(m)}\!=\!f_\theta(x_i^{(m)})$ and a joint embedding
$\mathbf{z}_i^{(TVA)}\!=\!f_\theta(T_i,V_i,A_i)$ from the same forward
backbone. \Cref{fig:method} overviews the three training objectives.

\subsection{Preliminary: Pairwise Alignment (\texorpdfstring{$\mathcal{L}_A$}{L\_A})}\label{sec:pairwise}
Existing open AVT
embedders~\cite{girdhar2023imagebind,zhu2024languagebind,tang2026wave,nvidia2025nemotron}
optimize a symmetric InfoNCE loss over modality pairs:
\begin{small}
\begin{align}
\mathcal{L}_{\text{NCE}}^{(m,m')}\!&=\!-\tfrac{1}{B}\!\sum_i\!\log
\tfrac{e^{\mathbf{z}_i^{(m)\!\top}\mathbf{z}_i^{(m')}/\tau}}
{\sum_j e^{\mathbf{z}_i^{(m)\!\top}\mathbf{z}_j^{(m')}/\tau}},
\label{eq:pairwise}\\
\mathcal{L}_A\!&=\!\!\!\sum_{(m,m')\in\mathcal{P}}\!\!\!
  \mathcal{L}_{\text{NCE}}^{(m,m')},\;
  \mathcal{P}\!=\!\{(T,V),(T,A),(V,A)\}.\nonumber
\end{align}
\end{small} The \textsc{Pairwise} baseline we report throughout uses
$\mathcal{L}_A$ alone. Since
$\mathcal{L}_A$ never operates on the joint $(T,V,A)$ vector,
$\mathbf{z}_{TVA}$ is neither supervised nor distilled: pairwise
alignment binds the three streams only to the extent that each
pair enforces in isolation. \method{} retains $\mathcal{L}_A$
and adds two joint-level losses, $\mathcal{L}_D$ and
$\mathcal{L}_T$, described next.

\subsection{Fusion-as-Teacher Distillation (\texorpdfstring{$\mathcal{L}_D$}{L\_D})}\label{sec:fat}

The joint forward of $f_\theta$ on $(T,V,A)$ is the only step
at which all three modalities interact in the network. We use
$\mathbf{z}_{TVA}$ as the teacher. A symmetric InfoNCE pulls
each single-modal embedding toward a stop-gradient copy of
$\mathbf{z}_{TVA}$ and pushes it away from the joint vectors
of other in-batch samples:
\begin{align}
\mathcal{L}_D \;=\; \tfrac{1}{|\mathcal{M}|}\sum_{m\in\mathcal{M}}
\mathcal{L}_{\text{NCE}}^{(m,\,\text{sg}(TVA))},
\label{eq:distill}
\end{align}
with $\mathcal{M}\!=\!\{T,V,A\}$. The teacher
$\mathrm{sg}(\mathbf{z}_{TVA})$ already encodes the three
modalities jointly, so each single-modal student is trained
against the joint geometry that an $A{+}T{+}V$ query reaches at
inference. Teacher and student share $f_\theta$ and live in the
same batch; the only additional cost over the pairwise baseline
is one extra joint forward pass.

\paragraph{Fusion vs.\ unimodal teacher.}
A unimodal teacher $g_m(x^{(m)})$, e.g.\
SigLIP~\cite{tschannen2025siglip2}, Whisper, or
BEATs~\cite{beats}, is trained on the marginal $p(x^{(m)})$ and
therefore carries no information about cross-modal neighbors
of $x$. By contrast, $\mathbf{z}_{TVA}$ is computed on the
joint input, so distillation into $\mathbf{z}_A$ propagates
$\partial\,\mathrm{sim}(\mathbf{z}_A,\mathrm{sg}(\mathbf{z}_{TVA}))/\partial\theta_A$,
which encodes the $T,V$ neighbors that an audio-only query
will see at inference. This predicts two effects, both of
which we confirm. Audio-anchored directions benefit most from
$\mathcal{L}_D$ (audio-related mean $+3.73$ vs.\ video-only
$+2.48$, \Cref{tab:supp:ablation:perdir}). 

\subsection{Tuple-InfoNCE Refinement (\texorpdfstring{$\mathcal{L}_T$}{L\_T})}\label{sec:tuple}
$\mathcal{L}_T$ is a regularizer that keeps
$\mathbf{z}_{TVA}$ informative about all three modalities,
preventing it from collapsing onto a dominant pair.  The joint vector $\mathbf{z}_{TVA}$ that $\mathcal{L}_D$ uses
as a teacher remains a passive average of three pair
geometries on top of pairwise alignment: $\mathcal{L}_A$ never
back-propagates through $\mathbf{z}_{TVA}$, so matched triples
are not pulled tighter than mismatched ones, and the
contribution of each modality is determined by whichever pair
gradient is strongest (in practice $T$–$V$). Audio is
under-represented in $\mathbf{z}_{TVA}$, a known
modality-imbalance effect in multimodal
fusion~\cite{wang2020balanced,peng2022balanced}.
$\mathcal{L}_T$ supervises $\mathbf{z}_{TVA}$ directly.

\paragraph{Tuple-InfoNCE.}
For a batch of size $B$ and modality set
$\mathcal{M}\!=\!\{T,V,A\}$, the joint similarity of an index pair
$(i,j)$ averages the six cross-modal cosines,
\begin{align}
s(i,j)\!=\!\tfrac{1}{M(M-1)}\!\!\sum_{m\neq m'}\!\!
\mathbf{z}_i^{(m)\!\top}\mathbf{z}_j^{(m')},
\label{eq:joint-sim}
\end{align}
and the Tuple-InfoNCE loss~\cite{liu2021contrastive,liu2020p4contrast} reads
\begin{align}
\mathcal{L}_T\!=\!-\tfrac{1}{B}\!\sum_i\!\log
\tfrac{e^{s(i,i)/\tau_T}}
{\sum_j e^{s(i,j)/\tau_T} + e^{s(i,\tilde i)/\tau_T}},
\label{eq:tuple-nce}
\end{align}
where $\tilde i$ is a modality-cycled hard negative described below.
When $M\!\geq\!3$, $s(i,j)$ is minimized over mismatch in
\emph{any} cross-modal direction, so $\mathcal{L}_T$ assigns the
largest gradient to the modality direction along which the matched
triple is currently slackest, providing the joint-level
supervision missing from pairwise alignment.

\paragraph{Modality-cycled hard negatives.}\label{sec:method:hn}
The in-batch grid in \Cref{eq:tuple-nce} contributes $B{-}1$
negatives, but each differs from the anchor in \emph{all three}
modalities, so its gradient pushes the joint cluster apart globally
without specifying which modality direction is too slack. We
therefore construct one targeted negative per anchor by shuffling
\emph{exactly one} modality of the batch with a derangement,
producing a tuple that disagrees with the anchor in a single slot;
the resulting contrastive gradient tightens the joint cluster along
that one direction. To prevent the joint geometry from drifting
back to a $T{+}V$-dominated configuration once any single modality
is tightened, we cycle the shuffled slot deterministically across
$\{T,V,A\}$ with period $3$, supervising every modality direction
in turn.

The three shuffled slots contribute asymmetrically. The
audio-shuffle slot produces the contrastive gradient least
redundant with the in-batch grid (audio carries the largest
caption gap in our training data). The text- and video-shuffle
slots prevent the joint cluster from drifting back to a
$T{+}V$-dominated geometry. We use $k\!=\!1$ per anchor; higher
$k$ overfits to modality-imbalanced
mismatches~\cite{hardneg2025}. $\mathcal{L}_D$ provides the
dominant single-loss gain, and $\mathcal{L}_T$ improves the
$A\!\leftrightarrow\!V$ routes on top of it
(\Cref{tab:supp:ablation:perdir}).

\subsection{Final Objective}
\label{sec:final-loss}

\method{} optimizes
\begin{align}
\mathcal{L}_{\method{}}\!=\!\lambda_D\mathcal{L}_D
+\lambda_T\mathcal{L}_T
+\lambda_A\mathcal{L}_A,
\label{eq:total}
\end{align}
with $(\lambda_D,\lambda_T,\lambda_A)\!=\!(1,1,1)$ chosen \emph{a
priori} (not tuned on \bench{}); \Cref{sec:supp:lossweight} shows that the recipe is
insensitive to the loss-weight ratio across a wide range. \Cref{alg:method} gives the per-step training
loop; the per-step cost over the pairwise baseline is one
extra joint forward pass, and inference latency is reported
in \Cref{sec:supp:compute}.

\begin{algorithm}[t]
\caption{\method{} training step (batch $\mathcal{B}$, step $t$).}
\label{alg:method}
\small
\begin{algorithmic}[1]
\State \textbf{Forward:} $\mathbf{z}^{(T)},\mathbf{z}^{(V)},\mathbf{z}^{(A)} \!\leftarrow\! f_\theta$ (3 single-modal); $\mathbf{z}^{(TVA)} \!\leftarrow\! f_\theta(T,V,A)$ (1 joint)
\State \textbf{Pairwise} $\mathcal{L}_A \leftarrow \sum_{(m,m')\in\{(T,V),(T,A),(V,A)\}} \mathcal{L}_{\text{NCE}}^{(m,m')}$
\State \textbf{Distill} $\mathcal{L}_D \leftarrow \tfrac{1}{3}\!\sum_{m\in\{T,V,A\}} \mathcal{L}_{\text{NCE}}^{(m,\,\mathrm{sg}(\mathbf{z}^{(TVA)}))}$
\State \textbf{Cycled HN:} $m_t \!\leftarrow\! \{T,V,A\}[t\bmod 3]$; draw derangement $\sigma$; build $\tilde{i}\!=\!(\dots,x_{\sigma(i)}^{(m_t)},\dots)$
\State \textbf{Tuple} $\mathcal{L}_T \!\leftarrow\! -\tfrac{1}{B}\!\sum_i \log\!\tfrac{e^{s(i,i)/\tau_T}}{\sum_j e^{s(i,j)/\tau_T} + e^{s(i,\tilde i)/\tau_T}}$
\State \textbf{Update} $\theta \leftarrow \theta - \eta\,\nabla_\theta(\lambda_D \mathcal{L}_D + \lambda_T \mathcal{L}_T + \lambda_A \mathcal{L}_A)$
\end{algorithmic}
\end{algorithm}

\section{Experiments}\label{sec:exp}

\subsection{Setup}\label{sec:setup}
\model{} is an adapter fine-tune of the open-weights
WAVE-7B~\cite{tang2026wave} backbone, with LoRA adapters on
the LLM trunk, an all-layer fusion head, and a BEATs adaptor;
full architectural details
are in \Cref{sec:supp:lora}.
The training data is a $1.5$\,M-triple
subset sampled from four public video--text datasets,
InternVid~\cite{wang2024internvid},
InternVid-FLT~\cite{wang2024internvid},
Panda-70M~\cite{chen2024panda70m}, and PVD~\cite{bolya2025pe},
together with a small in-house video corpus collected with
consent. We restrict the pool to clips that contain text,
video, and audio (\Cref{sec:supp:data}).
We release model weights, evaluation code, and \bench{}.
The training set is sample-identifier disjoint from every
evaluation pool in this paper, and the third-party benchmarks
(Clotho, SoundDescs, MSR-VTT, MSVD, DiDeMo, VATEX) are curated
by other groups under different caption styles.

We compare \model{}~({ours}) against three external
systems: the frozen open-weights backbone
{WAVE-7B}~\cite{tang2026wave}, the open
{Omni-Embed-Nemotron}~\cite{nvidia2025nemotron}, and closed {Gemini Embedding~2} via
Google's official endpoint with default settings. The endpoint ingests raw audio (WAV) and raw video
(MP4 with audio muxed) as inline data, so Gemini receives the
same clip bytes as \model{}. We cannot, however, inspect how
the closed system internally routes inline audio, so the
Gemini audio numbers reflect the currently deployed multipart
product rather than a model-capacity ceiling. Reported numbers
are the mean over three seeds $\{42, 43, 44\}$; aggregate seed
std is $\le 0.18$ R@1, which we treat as the noise floor.
The \textsc{Pairwise} ablation ($\mathcal{L}_A$ alone, the
recipe used by recent unified
embedders~\cite{wei2024uniir,jiang2025vlm2vec,zhang2024gme})
is reported in \Cref{tab:ablation}.

\paragraph{Systems we do not compare against.}
Qwen3-VL-Embedding~\cite{alibaba2026qwen3vl} has no audio path
and cannot serve the ten audio-anchored directions. A head-to-head comparison is left to
future work. The CLAP audio–text specialists
(\Cref{tab:external:audio}) have no video stream and cannot
serve the six $V$-anchored directions.

\begin{table*}[t]
\centering
\caption{\textbf{Comparison with prior cross-modal retrieval
benchmarks.} Each column under \emph{Single-modal} or
\emph{Dual-modal} marks whether the benchmark supports a given
retrieval direction-pair (a \cmark indicates both directions
of the pair are evaluated). \emph{Total} is the resulting
number of retrieval directions and $N$ is the size of the
standard test pool. \bench{} is the first public benchmark to
cover all three AVT modalities in one pool and the first to
evaluate dual-modal queries.}
\label{tab:bench-cmp}
\renewcommand{\arraystretch}{1.05}
\setlength{\tabcolsep}{6pt}
\small
\begin{tabular}{l c c c c c c r r}
\toprule
& \multicolumn{3}{c}{\textbf{Single-modal pairs}} & \multicolumn{3}{c}{\textbf{Dual-modal pairs}} & & \\
\cmidrule(lr){2-4}\cmidrule(lr){5-7}
\textbf{Benchmark}
 & $T\!\leftrightarrow\!V$
 & $T\!\leftrightarrow\!A$
 & $V\!\leftrightarrow\!A$
 & $T\!\leftrightarrow\!AV$
 & $A\!\leftrightarrow\!TV$
 & $V\!\leftrightarrow\!AT$
 & \textbf{Total} & $N$ \\
\midrule
\multicolumn{9}{l}{\emph{Video--text retrieval}} \\
MSR-VTT~\cite{xu2016msrvtt}                & \cmark & --     & --     & --     & --     & --     & 2 & $1{,}000$ \\
MSVD~\cite{chen2011msvd}                   & \cmark & --     & --     & --     & --     & --     & 2 & $670$ \\
DiDeMo~\cite{anne2017didemo}               & \cmark & --     & --     & --     & --     & --     & 2 & $1{,}004$ \\
VATEX~\cite{wang2019vatex}                 & \cmark & --     & --     & --     & --     & --     & 2 & $4{,}468$ \\
\midrule
\multicolumn{9}{l}{\emph{Audio--text retrieval}} \\
Clotho~\cite{drossos2020clotho}            & --     & \cmark & --     & --     & --     & --     & 2 & $1{,}045$ \\
SoundDescs~\cite{koepke2022audioretrieval} & --     & \cmark & --     & --     & --     & --     & 2 & $1{,}000$ \\
AudioCaps~\cite{kim2019audiocaps}          & --     & \cmark & --     & --     & --     & --     & 2 & $975$ \\
\midrule
\rowcolor{black!8}
\textbf{\bench{} (ours)}
 & \cmark & \cmark & \cmark & \cmark & \cmark & \cmark
 & \textbf{12} & \textbf{$3{,}782$} \\
\bottomrule
\end{tabular}
\end{table*}

\subsection{Results on Standard Audio--Text Benchmarks}
\label{sec:external:audio}

We evaluate \model{} zero-shot on two standard audio--text
retrieval benchmarks, Clotho~\cite{drossos2020clotho} and
SoundDescs~\cite{koepke2022audioretrieval}, and compare with
the published audio--text specialist literature, contemporary
open omni-modality embedders, and closed Gemini Embedding~2.
\Cref{tab:external:audio} reports R@1; full R@$k$ and NDCG@10
are in \Cref{tab:supp:external-audio}.

\begin{table*}[t]
\centering
\caption{\textbf{Audio--text retrieval R@1 on Clotho and
SoundDescs (zero-shot).} Specialised retrievers fine-tuned on
each task are marked FT. Cited rows draw on the original
papers' published values. \model{} outperforms closed Gemini
Embedding~2 by $13.3$ to $18.0$ R@1 in every direction; on
Clotho T$\to$A \model{}'s $19.14$ sits within $\sim$$2$ R@1 of
the current zero-shot contrastive SOTA band (Cacophony,
MGA-CLAP, M2D-CLAP at $20$ to $21$) without task-specific
fine-tuning.}
\label{tab:external:audio}
\renewcommand{\arraystretch}{1.0}
\setlength{\tabcolsep}{14pt}
\scriptsize
\begin{tabular}{lcc cc cc}
\toprule
& & & \multicolumn{2}{c}{\textbf{Clotho}} & \multicolumn{2}{c}{\textbf{SoundDescs}} \\
\cmidrule(lr){4-5}\cmidrule(lr){6-7}
\textbf{Model} & \textbf{Year} & \textbf{Setting} & \textbf{T$\to$A} & \textbf{A$\to$T} & \textbf{T$\to$A} & \textbf{A$\to$T} \\
\midrule
\multicolumn{7}{l}{\emph{Audio--text specialists}} \\
MMT~\cite{koepke2022audioretrieval}              & 2022 & FT & 6.7  & 7.0  & 30.7 & 31.4 \\
MS-CLAP                                          & 2023 & ZS & 16.7 & 20.0 & ---  & ---  \\
LAION-CLAP fusion~\cite{wu2023laionclap}         & 2023 & ZS & 18.2 & 25.7 & ---  & ---  \\
WavCaps-HTSAT~\cite{mei2023wavcaps}              & 2023 & ZS & 19.5 & 23.4 & ---  & ---  \\
WavCaps-CNN14~\cite{mei2023wavcaps}              & 2023 & ZS & \textbf{21.2} & 25.9 & --- & --- \\
FLAP fusion                                      & 2024 & ZS & 20.3 & 25.5 & ---  & ---  \\
Cacophony                                        & 2024 & ZS & 20.2 & 26.5 & ---  & ---  \\
MGA-CLAP                                         & 2024 & ZS & 20.8 & \textbf{26.5} & --- & --- \\
GLAP                                             & 2025 & ZS & 19.4 & 21.8 & ---  & ---  \\
M2D-CLAP~\cite{niizumi2025m2dclap}               & 2025 & ZS & 20.1 & 25.0 & ---  & ---  \\
\midrule
\multicolumn{7}{l}{\emph{Closed-source industrial baselines}} \\
Gemini Embedding~2                                   & 2026 & ZS & 5.2 & 1.3 & 7.0 & 7.4 \\
\midrule
\multicolumn{7}{l}{\emph{Open-source omni-modality embedders}} \\
ImageBind (ViT-H)~\cite{girdhar2023imagebind}        & 2023 & ZS & 6.0  & ---  & ---  & ---  \\
LanguageBind (CLIP-H/14)~\cite{zhu2024languagebind}  & 2024 & ZS & 16.7 & ---  & ---  & ---  \\
Omni-Embed-Nemotron~\cite{nvidia2025nemotron}        & 2025 & ZS & 6.4  & 3.5  & 6.4  & 4.8  \\
\rowcolor{black!8}
\textbf{\model{} (ours)}                             & 2026 & ZS & \textbf{19.1} & \textbf{16.1} & \textbf{25.0} & \textbf{20.7} \\
\bottomrule
\end{tabular}
\end{table*}

\model{} improves over closed Gemini Embedding~2 by $13.9$ to
$18.0$ R@1 on T$\to$A and $13.3$ to $14.7$ on A$\to$T.
Gemini's audio-anchored R@1 is consistently low across
benchmarks ($1.34$ on Clotho A$\to$T, $1.48$ on \bench{}
A$\to$T below), suggesting that its deployed multipart
endpoint does not route audio competitively for retrieval. The
CLAP-family specialists are audio--text dual-tower models
trained on AudioCaps-style pairs without video exposure, and
form the strongest single-task audio-retrieval band ($19$ to
$21$ R@1 on Clotho T$\to$A). \model{} reaches $19.14$ on
Clotho T$\to$A, within $\sim$$2$ R@1 of this band. On Clotho
A$\to$T, \model{} reaches $16.08$, still below the
specialist band ($25$ to $26.5$) but well above prior open
omni-modality systems on the same direction (ImageBind reports
$6.0$ T$\to$A, LanguageBind $16.7$ T$\to$A; neither reports
A$\to$T). On SoundDescs T$\to$A and A$\to$T, \model{} reaches
$25.00$ and $20.70$, the strongest open-system numbers in our
literature search. Scaling the training corpus is expected to push \model{}'s
A$\to$T performance toward and potentially beyond the
specialist band, since cross-modal supervision scales
naturally with additional multimodal data.

\subsection{Results on Standard Video--Text Benchmarks}
\label{sec:external:video}

We evaluate \model{} zero-shot on four standard video--text
benchmarks (MSR-VTT~\cite{xu2016msrvtt},
MSVD~\cite{chen2011msvd}, DiDeMo~\cite{anne2017didemo},
VATEX~\cite{wang2019vatex}) and compare with published
specialist literature and contemporary omni-modality
embedders.
\Cref{tab:external:video} reports R@1; full R@$k$ and NDCG@10
are in \Cref{tab:supp:external-video}.

\begin{table*}[t]
\centering
\caption{\textbf{Video--text retrieval R@1 on four standard
benchmarks (zero-shot).} Specialised video--text retrievers
fine-tuned per task are marked FT; the remainder are zero-shot.
Cited rows draw on each paper's published values. \model{}
matches the contemporary zero-shot specialist band on MSR-VTT
and MSVD and trails the strongest specialist with DiDeMo/VATEX
results, InternVideo2-1B, by $\sim$$9$ to $\sim$$12$ R@1 on
three of those four directions (and by $\sim$$30$ R@1 on the
exceptional VATEX V$\to$T). ``---'' denotes a number not
reported.}
\label{tab:external:video}
\renewcommand{\arraystretch}{1.0}
\setlength{\tabcolsep}{5pt}
\scriptsize
\resizebox{\linewidth}{!}{%
\begin{tabular}{ll cccc cccc}
\toprule
& & \multicolumn{4}{c}{\textbf{T$\to$V} (or T$\to$AV)} & \multicolumn{4}{c}{\textbf{V$\to$T} (or AV$\to$T)} \\
\cmidrule(lr){3-6}\cmidrule(lr){7-10}
\textbf{Model} & \textbf{Setting} & \textbf{MSR-VTT} & \textbf{MSVD} & \textbf{DiDeMo} & \textbf{VATEX} & \textbf{MSR-VTT} & \textbf{MSVD} & \textbf{DiDeMo} & \textbf{VATEX} \\
\midrule
\multicolumn{10}{l}{\emph{Specialised video--text retrievers (FT)}} \\
CLIP4Clip~\cite{luo2022clip4clip}      & FT & 44.5 & 46.2 & 43.4 & 55.9 & 42.7 & 56.0 & 42.5 & 73.4 \\
X-CLIP                                 & FT & 49.3 & 50.4 & 47.8 & ---  & 48.9 & 66.8 & 47.8 & ---  \\
\midrule
\multicolumn{10}{l}{\emph{Zero-shot video--text encoder band (specialists)}} \\
SigLIP-2 family~\cite{tschannen2025siglip2}    & ZS & 38.5--43.1 & 49.0--55.8 & --- & --- & 30.1--34.3 & 67.2--74.6 & --- & --- \\
PE-core family~\cite{bolya2025pe}              & ZS & 47.6--51.2 & 50.4--59.7 & --- & --- & 47.3--50.1 & 76.7--\textbf{85.4} & --- & --- \\
InternVideo2-1B~\cite{wang2024internvideo2}    & ZS & 51.9 & 58.1 & \textbf{57.0} & 70.4 & 50.9 & \textbf{83.3} & 54.3 & 85.4 \\
\midrule
\multicolumn{10}{l}{\emph{Closed-source industrial baselines}} \\
Gemini Embedding~2                                   & ZS & \textbf{53.9} & \textbf{77.1} & \textbf{55.6} & \textbf{69.4} & \textbf{48.3} & \textbf{77.9} & \textbf{53.3} & \textbf{66.7} \\
\midrule
\multicolumn{10}{l}{\emph{Open-source omni-modality embedders}} \\
ImageBind (ViT-H)~\cite{girdhar2023imagebind}        & ZS & 36.1 & ---  & ---  & ---  & ---  & ---  & ---  & ---  \\
LanguageBind (CLIP-H/14)~\cite{zhu2024languagebind}  & ZS & 44.8 & 53.9 & 39.9 & ---  & 40.9 & 72.0 & 39.8 & ---  \\
Omni-Embed-Nemotron~\cite{nvidia2025nemotron}        & ZS & 35.8 & 55.8 & 41.9 & 47.5 & 30.6 & 49.2 & 36.5 & 39.9 \\
\rowcolor{black!8}
\textbf{\model{} (ours)}                             & ZS & 47.6 & 66.9 & 46.0 & 58.0 & 43.7 & 63.3 & 45.1 & 55.0 \\
\bottomrule
\end{tabular}}
\end{table*}

\model{} reaches $47.6/43.7$ on MSR-VTT, matching
PE-coreB~\cite{bolya2025pe} ($47.6/47.3$) and improving over
SigLIP-2-L~\cite{tschannen2025siglip2}. On MSVD it reaches
$66.9/63.3$, improving over PE-coreL ($57.2$ T$\to$V). On
DiDeMo and VATEX, \model{} trails the strongest specialist
with results on both benchmarks,
InternVideo2-1B~\cite{wang2024internvideo2}, by $9$ to $12$
R@1 on three of the four directions and by $30$ R@1 on VATEX
V$\to$T, driven by InternVideo2's exceptional $85.4$ V$\to$T.
We attribute this gap to the data scale of single-task video
training; \model{} also has no audio specialisation for these
benchmarks, which carry no audio queries. \model{} improves
over LanguageBind on $5$ of $6$ reported video directions.
Closed Gemini wins all eight video directions by $5$ to $15$
R@1, consistent with reports that Gemini is trained on a
substantially larger closed video--text corpus; a data-parity
comparison is out of scope. Taken together with the audio results in
\Cref{tab:external:audio}, \model{} is the strongest open
unified omni-modal embedder in our literature search.

\paragraph{Why a unified AVT embedder.}
Pairing a CLAP-style audio specialist with an InternVideo2
video specialist covers four of the twelve directions, but
neither system produces embeddings for the six dual-modal
directions ($T\!\leftrightarrow\!AV$, etc.) in one shared
index, and neither can anchor an audio query against text
grounded in shared video context. \method{} targets the
setting in which a single shared $(T,V,A)$ index is required.

\model{} reduces the audio-capability gap among general-purpose
omni-modality embedders, reaches the zero-shot audio--text
specialist band on Clotho T$\to$A, and stays within reach of
single-task video--text specialists while covering all twelve
directions in one model.

\subsection{\bench{}: A 12-Direction AVT Probe}\label{sec:bench}

The benchmarks above evaluate single-modal directions only
and do not stress-test the joint representation
$\mathbf{z}_{TVA}$ that \method{} is built around. A
practical AVT retriever also needs to serve six dual-modal
queries ($T\!\leftrightarrow\!AV$, $A\!\leftrightarrow\!TV$,
$V\!\leftrightarrow\!AT$) that compose two modalities on one
side. No public suite covers these directions: standard
video--text and audio--text benchmarks each evaluate two
single-modal directions, never both video and audio in the
same pool.
\Cref{tab:bench-cmp} contrasts \bench{} with these prior
benchmarks across modality coverage, number of retrieval
directions, and evaluation-pool size.

\bench{} contains $3{,}782$ held-out aligned $(T,V,A)$
triples, scored against the same gallery in every direction;
the metric is Recall@1. Triples are curated from in-house
video sources that are disjoint from the training data at the
sample-identifier level. \textbf{All $3{,}782$ captions are
reviewed and corrected by trained human annotators}: each
caption starts from a Gemini 3.0 Pro draft, then the
annotator verifies alignment against the video and the audio
track and rewrites any inaccuracies.  Research-use licensing follows prior
released benchmarks built on proprietary
content~\cite{wang2019vatex,faysse2025colpali}; full
annotation guidelines, the distribution mismatch with
the training corpus, and a caption-paraphrase robustness
check are in
\Cref{sec:supp:bench,sec:supp:data,sec:supp:paraphrase}.

\Cref{tab:main} reports R@1 on all 12 directions of \bench{}.
\model{} reaches AVG-all $\mathbf{34.84}$, $+9.52$ over the
frozen WAVE-7B backbone, $+8.03$ over Omni-Embed-Nemotron, and
$+1.72$ over closed Gemini Embedding~2 ($33.12$). Both gaps
over external systems exceed the seed noise floor
($\le 0.18$ R@1 std). The gains concentrate on the eight
audio-anchored directions: $A\!\to\!T$, $\mathbf{11.92}$ vs.\
$1.48$ ($+10.44$); $A\!\to\!T{+}V$, $\mathbf{23.45}$ vs.\
$6.00$; $V\!\to\!A$, $\mathbf{25.46}$ vs.\ $13.80$. Gemini
wins the four visually-saturated $T\!\leftrightarrow\!V$ and
$T\!\leftrightarrow\!A{+}V$ routes by $6$ to $10$ R@1,
consistent with its larger closed video--text corpus.
A qualitative comparison is in \Cref{sec:supp:retrcmp}.

\begin{table*}[t]
\centering
\caption{\textbf{Recall@1 on the 12 directions of \bench{}.}
\textbf{Bold}: best per direction. The \textbf{ours} row is
shaded. The pairwise-only baseline (\textsc{Pairwise},
$\mathcal{L}_A$ alone) is reported in \Cref{tab:ablation}.}
\label{tab:main}
\renewcommand{\arraystretch}{1.18}
\setlength{\tabcolsep}{4.5pt}
\small
\resizebox{\linewidth}{!}{%
\begin{tabular}{@{}l cccccc cccccc ccc@{}}
\toprule
& \multicolumn{6}{c}{\textbf{Single-modal}}
& \multicolumn{6}{c}{\textbf{Dual-modal}}
& \multicolumn{3}{c}{\textbf{Averages}} \\
\cmidrule(lr){2-7}\cmidrule(lr){8-13}\cmidrule(lr){14-16}
\textbf{System}
& $t\!\to\!v$ & $v\!\to\!t$ & $t\!\to\!a$ & $a\!\to\!t$ & $v\!\to\!a$ & $a\!\to\!v$
& $t\!\to\!av$ & $av\!\to\!t$ & $a\!\to\!tv$ & $tv\!\to\!a$ & $v\!\to\!at$ & $at\!\to\!v$
& single & dual & all \\
\midrule
\multicolumn{16}{l}{\emph{Closed-source industrial baselines}} \\
Gemini Embedding~2
& \textbf{55.13} & \textbf{53.83} & 12.61 & 1.48 & 13.80 & 15.76
& \textbf{55.45} & \textbf{50.16} & 6.00 & 16.79 & \textbf{54.97} & \textbf{61.45}
& 25.44 & 40.80 & 33.12 \\
\midrule
\multicolumn{16}{l}{\emph{Open-source omni-modality embedders}} \\
WAVE-7B~\cite{tang2026wave}
& 41.27 & 34.45 &  8.67 &  3.38 & 14.89 & 12.93
& 44.95 & 42.68 &  2.27 & 15.65 & 38.74 & 43.92
& 19.27 & 31.37 & 25.32 \\
Omni-Embed-Nemotron~\cite{nvidia2025nemotron}
& 43.55 & 39.74 & 12.19 &  8.83 & 14.07 & 12.35
& 36.12 & 34.53 & 14.75 & 16.63 & 41.30 & 47.67
& 21.79 & 31.84 & 26.81 \\
\rowcolor{black!8}
\textbf{OmniRetriever (ours)}
& 48.89 & 46.30 & \textbf{14.20} & \textbf{11.92} & \textbf{25.46} & \textbf{24.99}
& 45.37 & 44.47 & \textbf{23.45} & \textbf{26.07} & 52.49 & 54.47
& \textbf{28.63} & \textbf{41.05} & \textbf{34.84} \\
\bottomrule
\end{tabular}}
\end{table*}

\subsection{Ablation and Analysis}
\label{sec:ablation}

\Cref{tab:ablation} reports the stair-step ablation against
the frozen WAVE-7B base on \bench{}, isolating the
contribution of each loss in the \method{} objective. The
per-direction breakdown is in \Cref{tab:supp:ablation:perdir}.

\paragraph{Pairwise baseline.}
The \textsc{Pairwise} baseline ($\mathcal{L}_A$ only, on our
training corpus) lifts AVG-all from $25.32$ to $31.08$
($+5.76$). The gain is unevenly distributed:
$T\!\leftrightarrow\!V$ and most dual-modal directions improve
by $5$ to $9$ R@1, but audio-anchored R@1 stays low
($a\!\to\!t$ $9.00$, $a\!\to\!tv$ $17.00$). The audio axis is
not capacity-bottlenecked here; it is bottlenecked by the
weak-fusion regime of pairwise alignment.

\paragraph{Fusion-as-teacher distillation ($\mathcal{L}_D$).}
Adding $\mathcal{L}_D$ lifts AVG-all by $+3.52$
($31.08\!\to\!34.60$), the dominant single-loss contribution.
This already improves on closed Gemini Embedding~2 by $+1.48$
R@1. The gain spans both regimes (video-only $+2.48$,
audio-related $+3.73$; \Cref{tab:supp:ablation:perdir}); the
audio side gains more because the joint forward routes $T,V$
context into the audio sub-encoder, which unimodal teachers
cannot do.

\paragraph{Tuple-InfoNCE refinement ($\mathcal{L}_T$).}
Adding $\mathcal{L}_T$ on top of $\mathcal{L}_D{+}\mathcal{L}_A$
lifts AVG-all by $+0.24$ to $\mathbf{34.84}$. The aggregate
gain is small but stable across seeds: the per-seed
$\Delta_{\mathcal{L}_T}$ is $\{+0.24, +0.21, +0.27\}$ on
$\{42, 43, 44\}$ (\Cref{tab:supp:seedstability}), well above
the aggregate seed std ($\le 0.07$). The per-direction breakdown
(\Cref{tab:supp:ablation:perdir}) shows that $\mathcal{L}_T$
redistributes capacity: it adds $+2.03$ to $+2.86$ R@1 on the
four $A\!\leftrightarrow\!V$ routes (above per-direction noise
of $\pm 0.5$ to $\pm 0.8$) while losing $\sim$$2$ R@1 on
text-anchored dual-modal routes. We keep $\mathcal{L}_T$ in
the released model because $A\!\leftrightarrow\!V$ is the
bottleneck direction for open AVT embedders.

\begin{table}[t]
\centering
\caption{\textbf{Ablation of the \method{} losses on \bench{}}
(AVG-all R@1; $\Delta$ vs.\ the frozen WAVE-7B base).
$\mathcal{L}_D$ on top of the \textsc{Pairwise} baseline adds
$+3.52$ R@1, already $+1.48$ over closed Gemini Embedding~2
($33.12$). $\mathcal{L}_T$ adds a further $+0.24$ on aggregate
and redistributes capacity toward $A\!\leftrightarrow\!V$
(\Cref{tab:supp:ablation:perdir}).}
\label{tab:ablation}
\renewcommand{\arraystretch}{1.0}
\scriptsize
\resizebox{\linewidth}{!}{%
\begin{tabular}{lcc}
\toprule
\textbf{Variant} & \textbf{AVG\,all} & $\Delta$ \\
\midrule
WAVE-7B (no fine-tune)                  & 25.32 & --- \\
\textsc{Pairwise} ($\mathcal{L}_A$ only)              & 31.08 & $+5.76$ \\
$\mathcal{L}_D{+}\mathcal{L}_A$ \;(no $\mathcal{L}_T$) & 34.60 & $+9.28$ \\
\textbf{\model{}}\,( $\mathcal{L}_T{+}\mathcal{L}_D{+}\mathcal{L}_A$) & \textbf{\underline{34.84}} & $\mathbf{+9.52}$ \\
\bottomrule
\end{tabular}}
\vspace{-8mm}
\end{table}

\paragraph{Discussion.}
Pairwise InfoNCE optimizes three pair softmaxes
independently, so the joint $(T,V,A)$ vector inherits whichever
pair carries the largest cosine gap (in practice $T$–$V$) and
leaves the audio axis slack. $\mathcal{L}_D$ supplies the
cross-modal teacher that closes the audio gap at the
single-modal level. $\mathcal{L}_T$ then redistributes the
remaining joint-level capacity toward the
$A\!\leftrightarrow\!V$ axis.  We conjecture that Gemini Embedding~2 follows a similar
pairwise contrastive recipe based on public reports\cite{google2026gemini2}; our \textsc{Pairwise} baseline
performs similarly to it in the same weak-fusion regime.

\section{Conclusion}\label{sec:conclusion}
A unified AVT encoder can computes a joint embedding
$\mathbf{z}_{TVA}$ on every $(T,V,A)$ forward, but pairwise
InfoNCE leaves it unsupervised. We turn $\mathbf{z}_{TVA}$
into a training signal: $\mathcal{L}_D$ distills its
stop-gradient copy into the single-modal embeddings, and
$\mathcal{L}_T$ supervises it directly against a tuple grid
with modality-cycled hard negatives. Both losses reuse the
same backbone forward and apply to any unified retriever
whose forward produces a joint embedding. A cross-backbone replication reproduces the dominant $\mathcal{L}_D$ effect.
\model{} reaches the zero-shot audio--text
specialist band on Clotho T$\to$A, reduces the open
omni-modality gap on A$\to$T, and improves over closed
Gemini Embedding~2 on \bench{} by $1.72$ AVG-all R@1. To probe the joint representation that pairwise training
cannot expose, we release \bench{}, the first AVT benchmark
scoring $3{,}782$ human-corrected  triples on a
shared gallery across all 12 single- and dual-modal
directions. Model weights, training code, and the benchmark
are released.

\clearpage
\section*{Limitations}
\label{sec:limitations}
\paragraph{Video--text gap.}
On the four $T\!\leftrightarrow\!V$ and
$T\!\leftrightarrow\!A{+}V$ directions, \model{} trails closed
Gemini Embedding~2 by $6$ to $10$ R@1 on \bench{} and by $5$
to $15$ R@1 on the external video benchmarks. We attribute
this gap to data scale: Gemini is trained on closed
video--text corpora reportedly orders of magnitude larger than
what is practically reachable in an academic AVT setting, and
a unified $(T,V,A)$ embedder at our training scale is not
expected to match dedicated video--text encoders on every
benchmark.

\paragraph{Closed-baseline comparability.}
Gemini Embedding~2 is accessed only through a deployed
multipart API. We feed it the same raw audio+video bytes
\model{} consumes, but we cannot inspect how the closed system
internally routes inline audio. The Gemini numbers therefore
reflect the deployed product rather than a model-capacity
ceiling.

\paragraph{Backbone coverage.}
\method{} is instantiated on a 7\,B WAVE backbone, with a
$3$\,B Omni-Embed-Nemotron replication in
\Cref{sec:supp:xbackbone}. Verification on other backbones
(e.g., Qwen3-VL-Embedding) is left to future work.

\paragraph{Compression.}
Our analyses focus on retrieval accuracy. We do not study
embedding compression beyond the post-hoc int8/binary
baselines reported in \Cref{sec:supp:compression}; a
compression-aware training recipe is left to future work.

\section*{Ethical Considerations}
\label{sec:ethics}
\paragraph{Privacy.}
Embedding inversion (vec2text~\cite{vec2text}) is a known
attack on retrieval embeddings. The training corpus is
sampled from four public video--text datasets (InternVid,
InternVid-FLT, Panda-70M, PVD) and a small in-house corpus
collected with consent; we use the public sources under the
terms of their original releases and do not redistribute the
assembled training subset. We recommend application-layer
defences such as DP-SGD or representation distortion for
deployments with sensitive content. \bench{} releases only
the source identifiers and caption text needed to reproduce
retrieval scores.

\paragraph{Intended use.}
We release \model{} and \bench{} for research on cross-modal
retrieval. The research-use license accompanying the release
prohibits deployment in surveillance applications affecting
natural persons.

\paragraph{Provenance and licensing.}
The training corpus combines four public video--text datasets
(InternVid, InternVid-FLT, Panda-70M, PVD) with a small
in-house corpus collected with consent. We do not redistribute
the assembled training subset. \bench{} is built from sources
held out from training and is released following the standard
video-benchmark protocol: source identifiers, clip intervals,
and captions are released, while the underlying media is not
redistributed. Curation details are in \Cref{sec:supp:bench}.

\bibliography{refs}

\appendix
\renewcommand{\thetable}{S\arabic{table}}
\renewcommand{\thefigure}{S\arabic{figure}}
\setcounter{table}{0}
\setcounter{figure}{0}

\section*{Appendix Overview}
\begin{itemize}[leftmargin=*,nosep,topsep=2pt]
  \item \Cref{sec:supp:lora,sec:supp:hp}: backbone, LoRA
        configuration, and full training hyper-parameters.
  \item \Cref{sec:supp:compute}: end-to-end inference latency
        on the four input paths.
  \item \Cref{sec:supp:lossweight}: sensitivity of \method{}
        to the loss weights
        $(\lambda_T,\lambda_D,\lambda_A)$.
  \item \Cref{sec:supp:data}: training-corpus description
        and caption-length statistics.
  \item \Cref{sec:supp:bench,sec:supp:duration}: \bench{}
        construction, licensing, and test-clip duration
        statistics.
  \item \Cref{sec:supp:compression}: post-hoc compression
        analysis (uniform dimension downsampling, int8 and
        binary quantization, front-dim truncation).
  \item \Cref{sec:supp:external}: per-$k$ R@1/5/10 and
        NDCG@10 on all six external benchmarks, with
        row-by-row literature comparison.
  \item \Cref{sec:supp:perdir-ablation}: stair-step
        per-direction decomposition of the
        $\mathcal{L}_A,\mathcal{L}_D,\mathcal{L}_T$
        contributions on \bench{}.
  \item \Cref{sec:supp:tupleorder}: triple-cosine intra/inter
        geometry on \bench{} across training stages.
  \item \Cref{sec:supp:xbackbone}: cross-backbone replication
        of \method{} on Omni-Embed-Nemotron-3B.
  \item \Cref{sec:supp:paraphrase}: caption-paraphrase
        robustness check on \bench{}.
  \item \Cref{sec:supp:seedstability}: per-seed $\mathcal{L}_T$
        ablation across three seeds.
  \item \Cref{sec:supp:retrcmp,sec:supp:fail}: qualitative
        retrieval comparisons and failure-case analysis.
\end{itemize}

\section{Backbone and LoRA Fine-Tuning Details}\label{sec:supp:lora}

We take \textbf{WAVE-7B}~\cite{tang2026wave}, an open-weights
extension of Qwen2.5-Omni with a hierarchical visual fusion and a
dual audio encoder, as our \emph{baseline backbone}. WAVE-7B is
the strongest open starting point on tri-modal tasks but, like all
prior open systems, relies on pairwise contrastive losses and lags
closed competitors on audio-involved retrieval; \method{} retrofits
the joint-objective recipe on top of it. The choice of WAVE-7B
(vs.\ alternative AVT backbones) is incidental to the \method{}
contribution: $\mathcal{L}_D$ and $\mathcal{L}_T$ act on the
backbone's joint and single-modal embeddings, not on its internal
architecture, and would transfer to any AVT model that exposes a
fusion path. We \emph{freeze} the entire WAVE-7B backbone, namely
the LLM trunk, the visual tower (SigLIP + merger), the audio tower
(Whisper), and the BEATs encoder transformer (all adopted from
WAVE without modification), and tune three small surfaces:
(i)~LoRA adapters
($r{=}16,\alpha{=}32$) inserted into the $q,k,v$ projections of
\emph{every} LLM layer (28 layers $\times$ 3 = 84 matrices,
$6.88$\,M params); (ii)~the all-layer fusion head
$\mathrm{Linear}(28d{\to}d)\!\to\!\mathrm{GELU}\!\to\!\mathrm{Linear}(d{\to}d)$
with $d{=}3584$, $372.51$\,M params; (iii)~the BEATs adaptor on top
of the frozen BEATs encoder (LayerNorm + projector, $15.61$\,M
params). The fusion head consumes the concatenation of the
last-token hidden states of all 28 LLM layers and projects them
into the shared $\mathbb{R}^{3584}$ embedding space; sub-modal
embeddings reuse the same head. Total trainable parameters:
$\approx$\,$395.0$\,M ($4.20\,\%$ of the $9.41$\,B backbone).
Training uses bf16 + DeepSpeed ZeRO-0, batch size 64, LR
$1\!\times\!10^{-5}$ cosine, 1 epoch over the training corpus,
on
$4\!\times\!$NVIDIA RTX PRO 6000 (Blackwell architecture, Max-Q
variant, $96$\,GB). Total wall-clock: $\approx$\,$109$\,h
($35.6$\,s/step over $11{,}192$ steps).

\section{Hyper-Parameters and Reproducibility}\label{sec:supp:hp}

\Cref{tab:supp:hp} lists every hyper-parameter that affects the
reported numbers. We run all reported numbers under three
independent seeds $\{42, 43, 44\}$ and report the mean;
aggregate seed std on \bench{} AVG-all is $\le 0.18$ R@1.

\begin{table*}[t]
\centering
\caption{\textbf{Full hyper-parameters for \method{} training.}
\textbf{Tunable surfaces:} LoRA on the q,k,v projections of all 28 LLM
layers (84 matrices) plus, at full precision, the all-layer fusion head
and the BEATs adaptor (LayerNorm + projector). \textbf{Frozen:} LLM trunk weights, vision tower (incl.\ merger),
Whisper audio tower, and the BEATs encoder backbone. \textbf{Trainable
budget:} $\approx$\,$395.0$\,M total ($4.20\%$ of the $9.41$\,B backbone),
distributed as $6.88$\,M (LoRA), $372.51$\,M (all-layer fusion head:
Linear($28d\!\to\!d$)+GELU+Linear($d\!\to\!d$) with $d\!=\!3584$), and
$15.61$\,M (BEATs adaptor). \textbf{Loss settings:} single shared
temperature $\tau$ (no per-pair $\tau_{mm'}$); hard negatives are
deterministically cycled $\{$T-shuffle, A-shuffle, V-shuffle$\}$ with
period $3$, tightening the joint cluster on every modality
direction (\Cref{sec:method:hn} in main paper).}
\label{tab:supp:hp}
\renewcommand{\arraystretch}{1.15}
\small
\begin{tabularx}{\linewidth}{@{}p{0.35\linewidth} X@{}}
\toprule
\textbf{Hyper-parameter} & \textbf{Value} \\
\midrule
Backbone (baseline)             & WAVE-7B~\cite{tang2026wave} \\
LoRA rank $r$ / $\alpha$ / dropout & 16 / 32 / 0.05 \\
Embedding dim $d$               & 3584 \\
Text max length                 & 2048 tokens \\
Video frames / Audio duration   & 8 / 8\,s @ 16\,kHz \\
Audio encoders / Vision teacher & Whisper-Large-v3 + BEATs / SigLIP-2-SO400m \\
Optimiser                       & AdamW ($\beta_1{=}0.9, \beta_2{=}0.95$) \\
Learning rate / Warm-up / WD    & $10^{-5}$ cosine / $3\%$ / $0.01$ \\
Batch size                      & 64 (8/GPU $\times$ 4 GPUs $\times$ 2 acc) \\
Epochs / Precision / DeepSpeed  & 1 / bf16 / ZeRO-0 \\
Tuple-InfoNCE temperature $\tau$           & 0.01 \\
Loss weights $(\lambda_T,\lambda_D,\lambda_A)$ & $(1.0, 1.0, 1.0)$ \\
Hardware                        & 4$\times$RTX PRO 6000 Blackwell-MaxQ \\
Total wall-clock (11{,}192 steps) & $\approx$\,$109$\,h ($35.6$\,s/step) \\
\bottomrule
\end{tabularx}
\end{table*}

\section{Inference Latency}\label{sec:supp:compute}
\Cref{tab:supp:e2e-latency} reports full-stack single-query
latency on one RTX PRO 6000 Blackwell Max-Q GPU at bf16, batch
$=1$, averaged over $10$ timed iterations after $3$ warmups.
The measured forward pass includes the vision and audio towers
(when applicable), the LM trunk with all-layer hidden states
retained, and the all-layer fusion head
($28d\!\to\!d\!\to\!d$). The text-only path skips both towers
and is the cheapest. The joint AV path consumes the longest
token sequence ($\sim$$470$ tokens) and is the most expensive.
Peak GPU memory across all four paths is $20.13$\,GB; model
load alone uses $17.92$\,GB.

\begin{table}[!ht]
\centering
\caption{\textbf{End-to-end inference latency of \model{}.}
Mean$\pm$std over $10$ iterations after $3$ warmups; bf16,
batch $=1$, RTX PRO 6000 Blackwell Max-Q. Text uses $24$ tokens,
video uses $8$ frames at $224$\,px ($\sim$$268$ LM tokens), audio
uses $8$\,s of $16$\,kHz mono, joint AV concatenates
video + audio + prompt ($\sim$$470$ tokens). Cosine search and
text-side preprocessing are not included.}
\label{tab:supp:e2e-latency}
\small
\setlength{\tabcolsep}{8pt}
\begin{tabular}{lr}
\toprule
\textbf{Inference path} & \textbf{Latency (ms)} \\
\midrule
Text-only       & $\phantom{0}20.9\,\pm\,0.1$ \\
Video-only      & $105.6\,\pm\,0.2$ \\
Audio-only      & $\phantom{0}34.6\,\pm\,0.1$ \\
Joint AV        & $125.1\,\pm\,0.1$ \\
\bottomrule
\end{tabular}
\end{table}

The LM trunk on the merged token sequence dominates the cost.
A video-only forward spends roughly two thirds of its
$105.6$\,ms in the LM trunk, after the vision tower outputs
$\sim$$256$ LM-dim tokens. The joint AV path runs $125$\,ms
because the LM trunk now processes $\sim$$470$ tokens under
the same fp32 fusion head. Text-only is the practical floor at
$20.9$\,ms. In production, the embedding step is computed once
per query and the gallery search dominates per-query latency;
that search cost depends on embedding dim and dtype
(\Cref{sec:supp:compression}) rather than backbone compute.

\section{Loss-Weight Sensitivity}\label{sec:supp:lossweight}

The released \model{} uses uniform loss weights
$(\lambda_T,\lambda_D,\lambda_A)\!=\!(1,1,1)$. To test
sensitivity to this choice, we re-train \method{} from the
same WAVE-7B initialisation under the skewed setting
$(\lambda_T,\lambda_D,\lambda_A)\!=\!(50,10,1)$, which scales
the Tuple-InfoNCE loss by $50\times$ and the distillation loss
by $10\times$ relative to pairwise alignment.
\Cref{tab:supp:lossweight} reports both configurations on the
$3{,}782$-sample \bench{}.

\begin{table*}[t]
\centering
\caption{\textbf{Loss-weight sensitivity on \bench{} R@1.}
\model{} under two loss-weight configurations
$(\lambda_T,\lambda_D,\lambda_A)$. Both configurations sit
above closed Gemini Embedding~2 ($33.12$) and the pairwise-CL
ablation ($31.08$) on AVG-all. Differences between the two
settings are within $\le\!2$ R@1 in every direction and within
$0.71$ R@1 on AVG-all. We use the uniform $(1,1,1)$ setting for
the released model.}
\label{tab:supp:lossweight}
\renewcommand{\arraystretch}{1.15}
\setlength{\tabcolsep}{14pt}
\small
\begin{tabular}{lrr r}
\toprule
\textbf{Direction} & $(1,1,1)$ & $(50,10,1)$ & $\Delta$ \\
\midrule
text$\rightarrow$video        & 48.89 & 47.41 & $-$1.48 \\
video$\rightarrow$text        & 46.30 & 45.08 & $-$1.22 \\
text$\rightarrow$audio        & 14.20 & 13.78 & $-$0.42 \\
audio$\rightarrow$text        & 11.92 & 11.42 & $-$0.50 \\
video$\rightarrow$audio       & 25.46 & 25.12 & $-$0.34 \\
audio$\rightarrow$video       & 24.99 & 25.73 & $+$0.74 \\
text$\rightarrow$audio+video  & 45.37 & 43.42 & $-$1.95 \\
audio+video$\rightarrow$text  & 44.47 & 42.20 & $-$2.27 \\
audio$\rightarrow$text+video  & 23.45 & 24.64 & $+$1.19 \\
text+video$\rightarrow$audio  & 26.07 & 25.78 & $-$0.29 \\
video$\rightarrow$audio+text  & 52.49 & 50.45 & $-$2.04 \\
audio+text$\rightarrow$video  & 54.47 & 54.52 & $+$0.05 \\
\midrule
\textbf{AVG\,single}          & \textbf{28.63} & 28.09 & $-$0.54 \\
\textbf{AVG\,dual}            & \textbf{41.05} & 40.17 & $-$0.88 \\
\textbf{AVG\,all}             & \textbf{34.84} & 34.13 & $-$0.71 \\
\bottomrule
\end{tabular}
\end{table*}

\paragraph{Discussion.}
Both configurations sit comfortably above the closed Gemini
baseline ($33.12$) and the pairwise-CL ablation ($31.08$), so
the method's central claim is unaffected by the specific
weighting. The heavily-anchor-weighted $(50,10,1)$ setting
slightly improves audio-anchored directions ($a\!\to\!v$ $+0.74$,
$a\!\to\!tv$ $+1.19$) at the cost of text-anchored ones
($t\!\to\!av$ $-1.95$, $av\!\to\!t$ $-2.27$): putting more loss
mass on the fusion anchor pulls the audio sub-encoder closer to
it but de-emphasizes the pairwise alignment that benefits
text-anchored composition. The uniform setting is a
Pareto-friendly default. Per-pair temperature tuning or curriculum
schedules over $(\lambda_T,\lambda_D,\lambda_A)$ are left as future work.


\section{Training Corpus Description}\label{sec:supp:data}

\paragraph{Sources and scale.}
The training corpus contains approximately $1.5$\,M
$(T,V,A)$ triples sampled from four public video--text
datasets, InternVid~\cite{wang2024internvid},
InternVid-FLT~\cite{wang2024internvid},
Panda-70M~\cite{chen2024panda70m}, and PVD~\cite{bolya2025pe},
plus a small in-house video corpus collected with consent.
We restrict the pool to clips that contain all three
modalities: a text caption, video frames, and an audible
audio track. We do not redistribute the assembled training
subset; the public sources can be downloaded from their
original releases. \bench{} is built from sources held out
from training, with sample-identifier disjointness enforced.

\paragraph{Caption sources.}
For triples drawn from the public datasets, we keep the
captions provided by the original release. For the in-house
portion, captions are produced by an automatic multimodal
captioning pipeline with a held-out coherence filter on
$(T,V,A)$ alignment. \bench{} captions, in contrast, are
human-corrected: each caption begins as a Gemini 3.0 Pro
draft and is then verified against the video and audio and
rewritten where necessary by a trained annotator. The
resulting \bench{} captions therefore reflect a distinct
human-validated distribution from any caption in the training
corpus. We additionally test sensitivity to caption surface
form with the paraphrase analysis in
\Cref{sec:supp:paraphrase}.

\paragraph{Caption length distribution.}
\Cref{tab:supp:capstats} reports caption-word quantiles
measured on the training corpus. Captions are dense and
uniform, with mean $32.0$ words, median $30$, and
$99$\textsuperscript{th} percentile $76$.

\begin{table*}[t]
\centering
\caption{\textbf{Caption-word statistics of the training corpus.}}
\label{tab:supp:capstats}
\small
\renewcommand{\arraystretch}{1.15}
\setlength{\tabcolsep}{16pt}
\begin{tabular}{lrrrrrr}
\toprule
 & mean & median & p90 & p99 & max & min \\
\midrule
Caption (\#words) & 32.04 & 30 & 49 & 76 & 11{,}938 & 1 \\
Prompt (\#words)  &  5.00 &  5 &  5 &  5 &      5 & 5 \\
\bottomrule
\end{tabular}
\end{table*}

\section{\bench{}: Construction and License}\label{sec:supp:bench}

The \bench{} evaluation pool contains $3{,}782$ held-out
$(T,V,A)$ triples; the same pool serves as queries \emph{and}
gallery for all twelve retrieval directions. \bench{} is
released for research use under a custom license that mirrors
prior released benchmarks built on proprietary video
content~\cite{wang2019vatex,faysse2025colpali}: caption text,
public source identifiers, and retrieval indices are
redistributed, while the underlying media is not. \bench{}
triples come from internal video sources held out from the
training subset, with sample-identifier disjointness from the
training data enforced. \Cref{fig:supp:bench-grid} visualises
the 12 valid query$\to$target cells of the $6\!\times\!6$
source/target matrix.

\begin{figure}[t]
\centering
\resizebox{\columnwidth}{!}{%
\begin{tikzpicture}[
    every node/.style={font=\scriptsize},
    cell/.style={draw=black!50,line width=0.4pt,minimum width=1.05cm,
                 minimum height=0.55cm,inner sep=0pt,fill=white,align=center},
    sng/.style={cell,fill=black!10},
    dual/.style={cell,fill=black!25},
    diag/.style={cell,fill=black!80,text=white,font=\scriptsize\bfseries},
    hd/.style={font=\scriptsize\bfseries},
    block/.style={font=\tiny\itshape,text=black!55}
  ]
  \foreach \x/\t in {1/T,2/V,3/A,4/AV,5/AT,6/TV}{
    \node[hd] at (\x*1.05, 0.55) {\t};
  }
  \foreach \y/\t in {1/T,2/V,3/A,4/AV,5/AT,6/TV}{
    \node[hd] at (0.0, -\y*0.55) {\t};
  }
  \node[hd,anchor=south,font=\scriptsize\bfseries] at (3.7, 1.05)
       {target $\to$};
  \node[hd,anchor=east,rotate=90,font=\scriptsize\bfseries] at (-0.6, -1.93)
       {query $\to$};
  \foreach \src/\sy in {T/1, V/2, A/3}{
    \foreach \tgt/\tx in {T/1, V/2, A/3}{
      \ifx\src\tgt
        \node[diag] at (\tx*1.05, -\sy*0.55) {---};
      \else
        \node[sng] at (\tx*1.05, -\sy*0.55) {\src$\to$\tgt};
      \fi
    }
  }
  \node[dual] at (4*1.05, -1*0.55) {T->AV};
  \node[diag] at (5*1.05, -1*0.55) {---};
  \node[diag] at (6*1.05, -1*0.55) {---};
  \node[diag] at (4*1.05, -2*0.55) {---};
  \node[dual] at (5*1.05, -2*0.55) {V->AT};
  \node[diag] at (6*1.05, -2*0.55) {---};
  \node[diag] at (4*1.05, -3*0.55) {---};
  \node[diag] at (5*1.05, -3*0.55) {---};
  \node[dual] at (6*1.05, -3*0.55) {A->TV};
  \node[dual] at (1*1.05, -4*0.55) {AV->T};
  \node[diag] at (2*1.05, -4*0.55) {---};
  \node[diag] at (3*1.05, -4*0.55) {---};
  \node[diag] at (1*1.05, -5*0.55) {---};
  \node[dual] at (2*1.05, -5*0.55) {AT->V};
  \node[diag] at (3*1.05, -5*0.55) {---};
  \node[diag] at (1*1.05, -6*0.55) {---};
  \node[diag] at (2*1.05, -6*0.55) {---};
  \node[dual] at (3*1.05, -6*0.55) {TV->A};
  \foreach \tx in {4,5,6}{
    \foreach \sy in {4,5,6}{
      \node[diag] at (\tx*1.05, -\sy*0.55) {---};
    }
  }
\end{tikzpicture}}
\caption{\textbf{\bench{} retrieval directions.} 12 valid
query$\to$target cells across a $6\!\times\!6$ matrix: 6
single$\leftrightarrow$single (top-left $3\!\times\!3$), 3
single$\to$dual (top-right), 3 dual$\to$single (bottom-left). Black
cells are excluded by the task definition (same-modality matches
and dual$\to$dual).}
\label{fig:supp:bench-grid}
\end{figure}

\begin{figure*}[t]
\centering
\includegraphics[width=\linewidth]{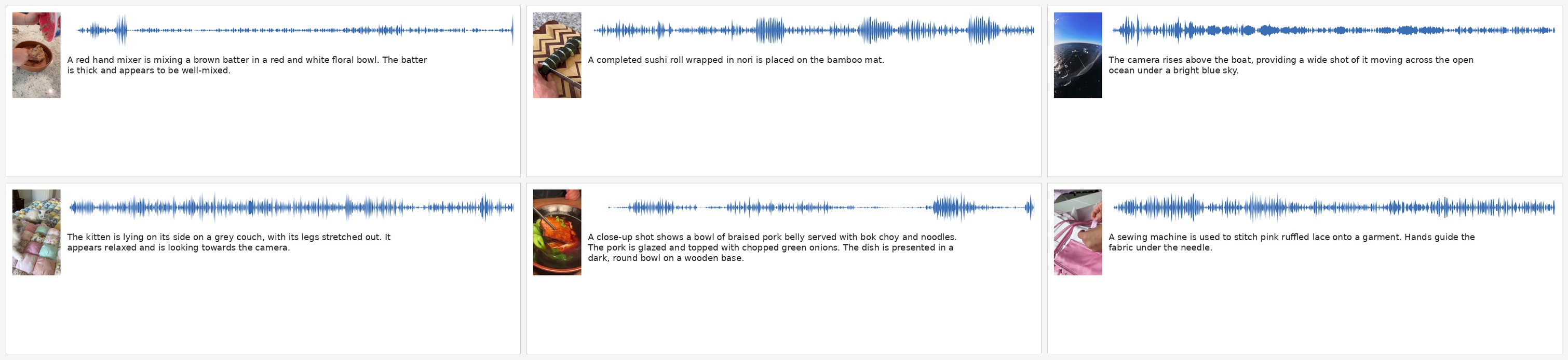}
\caption{\textbf{\bench{} sample cards (6 of $3{,}782$ held-out
triples).} Each card pairs a mid-clip video keyframe and the
recorded audio waveform with the caption used as $T$. The
displayed samples are everyday user-generated content covering
cooking, scenery, pets, narration over still imagery, ambient
music, and casual dialogue, representative of the natural mix
of short-form video in the pool.}
\label{fig:bench-samples}
\end{figure*}

\paragraph{Distribution.}
Following the protocol of prior video-language
benchmarks~\cite{wang2019vatex,xu2016msrvtt,chen2011msvd},
we redistribute the \bench{} evaluation pipeline, the
held-out caption text, source video identifiers, and clip
time intervals; the underlying media is not redistributed,
and users obtain it from the source platform via the released
identifiers. Sample-identifier disjointness with the
training subset is enforced via SHA-256 identifier comparison.

\section{Extended Compression Analysis}\label{sec:supp:compression}

\Cref{tab:supp:compression-12dir} reports the per-direction
breakdown under three operating points: the 3584-d fp32
default, a 1024-d int8 setting (uniform dimension downsampling
followed by symmetric per-vector min-max int8 quantization),
and a 512-d binary setting (uniform downsampling followed by
sign hashing). The same random indices are applied to query
and gallery, and we report the mean over $5$ random seeds. All
compression is applied post-hoc on the released \model{}
embeddings and on Gemini's API output without retraining, so
the reported numbers lower-bound what an MRL/QAT-trained head
would achieve. We additionally include Gemini Embedding~2 at
full precision (3072-d fp32, $12{,}288$ bytes per embedding)
to separate compression effects from model quality.

\begin{table*}[t]
\centering
\caption{\textbf{Per-direction R@1 of \model{} and Gemini under
post-hoc uniform downsampling and quantization} (mean over $5$
random-dim seeds, std $\le 0.6$ everywhere). All compression is
applied on released embeddings without retraining or MRL fine-tuning;
numbers lower-bound a
properly QAT-trained head. \textbf{Storage:} fp32 = $14{,}336$\,B
(\model{}) / $12{,}288$\,B (Gemini); int8/1024 = $1{,}024$\,B; bin/512 =
$64$\,B. Even at int8/1024 \model{}'s audio-anchored gains are preserved
(\eg{}, audio$\to$text $5.43$ vs.\ Gemini's $1.48$ at full precision;
audio$\to$text+video $21.18$ vs.\ $6.00$).}
\label{tab:supp:compression-12dir}
\renewcommand{\arraystretch}{1.15}
\setlength{\tabcolsep}{12pt}
\small
\begin{tabular}{lrrrrrr}
\toprule
\textbf{Direction} &
\multicolumn{3}{c}{\textbf{\model{}}} &
\multicolumn{3}{c}{\textbf{Gemini Emb-2}} \\
\cmidrule(lr){2-4}\cmidrule(lr){5-7}
& fp32 & int8/1024 & bin/512 & fp32 & int8/1024 & bin/512 \\
\midrule
text$\rightarrow$video       & 48.89 & 34.87 &  4.12 & 55.13 & 52.43 & 38.39 \\
video$\rightarrow$text       & 46.30 & 27.44 &  2.03 & 53.83 & 48.25 & 37.69 \\
text$\rightarrow$audio       & 14.20 &  8.60 &  0.91 & 12.61 & 11.99 &  7.69 \\
audio$\rightarrow$text       & 11.92 &  5.43 &  0.45 &  1.48 &  1.38 &  6.37 \\
video$\rightarrow$audio      & 25.46 & 23.11 & 11.46 & 13.80 & 11.64 &  8.58 \\
audio$\rightarrow$video      & 24.99 & 22.94 & 11.59 & 15.76 & 13.44 &  7.89 \\
text$\rightarrow$audio+video & 45.37 & 30.29 &  3.93 & 55.45 & 52.60 & 35.56 \\
audio+video$\rightarrow$text & 44.47 & 27.98 &  2.91 & 50.16 & 43.56 & 34.53 \\
audio$\rightarrow$text+video & 23.45 & 21.18 &  8.63 &  6.00 &  5.57 &  9.15 \\
text+video$\rightarrow$audio & 26.07 & 24.30 & 10.40 & 16.79 & 15.04 &  9.78 \\
video$\rightarrow$audio+text & 52.49 & 48.47 & 24.81 & 54.97 & 50.31 & 41.15 \\
audio+text$\rightarrow$video & 54.47 & 51.72 & 28.47 & 61.45 & 58.27 & 42.74 \\
\midrule
\textbf{AVG\,single} & 28.63 & 20.40 &  5.09 & 25.44 & 23.19 & 17.77 \\
\textbf{AVG\,dual}   & 41.05 & 33.99 & 13.19 & 40.80 & 37.56 & 28.82 \\
\textbf{AVG\,all}    & \textbf{34.84} & 27.19 &  9.14 & 33.12 & 30.37 & 23.29 \\
\bottomrule
\end{tabular}
\end{table*}

At full precision \model{} ($34.84$) and Gemini Embedding~2
($33.12$) are within $\sim$$1.7$ R@1; under uniform-downsample
$14\!\times$ byte compression to int8/1024 \model{} drops to
$27.19$ vs.\ Gemini's $30.37$, and under $224\!\times$ binary
compression to $9.14$ vs.\ Gemini's $23.29$. The asymmetric
degradation indicates that Gemini's embedding space is
significantly more isotropic than ours, consistent with Gemini
having been trained for compression while \model{} has not.
At int8/1024 \model{}'s audio-anchored gains are still
preserved (\eg{} bench $A\!\to\!T$ $5.43$ vs.\ Gemini's $1.38$;
$A\!\to\!T{+}V$ $21.18$ vs.\ $5.57$), so the audio-capability
finding is robust to a $14\!\times$ byte-compression budget.
Gemini's bench $A\!\to\!T$ rises from $1.48$ to $6.37$ under
binary compression, which looks like a paradox. The cause is
$\{-1,+1\}$ sign-hashing collapsing a near-degenerate audio
embedding subspace onto a small set of buckets, which inflates
random-collision recall on a $\sim$$4$K gallery; the behaviour
is consistent across the $5$ random-dim seeds (std $\le 0.6$).
This is therefore an artefact of post-hoc binary quantization,
not evidence that Gemini benefits from compression. We leave compression-aware training (MRL+QAT) for
future work.

\paragraph{Front-dim truncation sanity check.}
As an alternative to random downsampling, we also keep the first
$d$ dimensions of the embedding (the naive ``head-of-vector''
fallback used by some MRL deployments) and re-apply the same
int8/binary quantization. Differences vs.\ random downsampling are
small ($+0.62$ R@1 at int8/1024, $+0.21$ at bin/512 on AVG-all,
within $1$ to $2\sigma$ of the random-sample spread), so the two
schemes are substantively equivalent. We adopt the random-sample
formulation throughout since it does not assume any particular
basis ordering.

\section{\bench{} Test-Clip Duration Statistics}\label{sec:supp:duration}

\Cref{tab:supp:duration} shows duration quantiles measured on the
\bench{} test pool ($N\!=\!3{,}782$). The quantiles justify our
32-frame / 0.5\,s video sampling and 8\,s audio crop at inference time.

\begin{table*}[t]
\centering
\caption{\textbf{Duration quantiles (seconds) on the \bench{} test pool.}
($N\!=\!3{,}782$). Audio is the original recorded waveform.}
\label{tab:supp:duration}
\small
\renewcommand{\arraystretch}{1.15}
\setlength{\tabcolsep}{16pt}
\resizebox{\linewidth}{!}{%
\begin{tabular}{lrrrrrr}
\toprule
\textbf{Stream} & mean & median & p10 & p90 & p99 & max \\
\midrule
Video & 3.25 & 2.16 & 1.10 & 6.13 & 16.17 & 30.16 \\
Audio & 3.13 & 2.00 & 0.98 & 5.99 & 16.02 & 30.00 \\
\bottomrule
\end{tabular}}
\end{table*}

\section{External Cross-Modal Retrieval Benchmarks}\label{sec:supp:external}

This section complements main paper \Cref{sec:external:audio,sec:external:video}
(\Cref{tab:external:audio,tab:external:video}) with (a) the full
row-by-row specialist literature on Clotho audio--text retrieval
(\Cref{tab:supp:external-audio-full}), (b) the row-by-row video--text
SOTA placement with each size variant of SigLIP-2 / PE-core broken
out (\Cref{tab:supp:external-video-rows}), and (c) per-$k$ and
NDCG@10 details for all six benchmarks
(\Cref{tab:supp:external-video,tab:supp:external-audio}).
The main paper consolidates specialist rows into year-grouped bands;
the full row-by-row versions below preserve all individual numbers
for readers who want to look up specific systems.

\begin{table}[t]
\centering
\caption{\textbf{Full Clotho and SoundDescs audio--text retrieval R@1
(zero-shot).} Row-by-row specialist literature, expanded version of
main-paper \Cref{tab:external:audio}.}
\label{tab:supp:external-audio-full}
\renewcommand{\arraystretch}{1.1}
\setlength{\tabcolsep}{5pt}
\small
\resizebox{\linewidth}{!}{%
\begin{tabular}{lcc cc cc}
\toprule
& & & \multicolumn{2}{c}{\textbf{Clotho}} & \multicolumn{2}{c}{\textbf{SoundDescs}} \\
\cmidrule(lr){4-5}\cmidrule(lr){6-7}
\textbf{Model} & \textbf{Year} & \textbf{Setting} & \textbf{T$\to$A} & \textbf{A$\to$T} & \textbf{T$\to$A} & \textbf{A$\to$T} \\
\midrule
MMT~\cite{koepke2022audioretrieval}    & 2022 & FT & 6.7   & 7.0   & 30.7  & 31.4 \\
ImageBind (ViT-H)~\cite{girdhar2023imagebind}      & 2023 & ZS & 6.0   & ---   & ---   & ---  \\
MS-CLAP                                & 2023 & ZS & 16.7  & 20.0  & ---   & ---  \\
LAION-CLAP fusion~\cite{wu2023laionclap}    & 2023 & ZS & 18.2  & 25.7  & ---   & ---  \\
WavCaps-HTSAT~\cite{mei2023wavcaps}    & 2023 & ZS & 19.5  & 23.4  & ---   & ---  \\
WavCaps-CNN14~\cite{mei2023wavcaps}    & 2023 & ZS & \textbf{21.2}  & 25.9  & ---   & ---  \\
LanguageBind (CLIP-H/14)~\cite{zhu2024languagebind} & 2024 & ZS & 16.7 & --- & --- & --- \\
FLAP fusion                            & 2024 & ZS & 20.3  & 25.5  & ---   & ---  \\
Cacophony                              & 2024 & ZS & 20.2  & 26.5  & ---   & ---  \\
MGA-CLAP                               & 2024 & ZS & 20.8  & \textbf{26.5}  & ---   & ---  \\
GLAP                                   & 2025 & ZS & 19.4  & 21.8  & ---   & ---  \\
M2D-CLAP~\cite{niizumi2025m2dclap}     & 2025 & ZS & 20.1  & 25.0  & ---   & ---  \\
\midrule
Gemini Embedding 2 (closed)            & 2026 & ZS & 5.19  & 1.34  & 7.00  & 7.37 \\
\rowcolor{black!6}
\textbf{\model{} (ours)}               & 2026 & ZS & \textbf{19.14} & 16.08 & \textbf{25.00} & \textbf{20.70} \\
\bottomrule
\end{tabular}}
\end{table}

\begin{table*}[t]
\centering
\caption{\textbf{Full video--text retrieval R@1 (zero-shot).}
Row-by-row, with SigLIP-2 and PE-core variants broken out, expanded
version of main-paper \Cref{tab:external:video}.}
\label{tab:supp:external-video-rows}
\renewcommand{\arraystretch}{1.1}
\setlength{\tabcolsep}{6pt}
\small
\resizebox{\linewidth}{!}{%
\begin{tabular}{ll cccc cccc}
\toprule
& & \multicolumn{4}{c}{\textbf{T$\to$V} (or T$\to$AV)} & \multicolumn{4}{c}{\textbf{V$\to$T} (or AV$\to$T)} \\
\cmidrule(lr){3-6}\cmidrule(lr){7-10}
\textbf{Model} & \textbf{Setting} & \textbf{MSR-VTT} & \textbf{MSVD} & \textbf{DiDeMo} & \textbf{VATEX} & \textbf{MSR-VTT} & \textbf{MSVD} & \textbf{DiDeMo} & \textbf{VATEX} \\
\midrule
CLIP4Clip~\cite{luo2022clip4clip}      & FT & 44.5 & 46.2 & 43.4 & 55.9 & 42.7 & 56.0 & 42.5 & 73.4 \\
X-CLIP~\cite{ma2022x}                                 & FT & 49.3 & 50.4 & 47.8 & ---  & 48.9 & 66.8 & 47.8 & ---  \\
ImageBind (ViT-H)~\cite{girdhar2023imagebind}    & ZS & 36.1 & ---  & ---  & ---  & ---  & ---  & ---  & ---  \\
SigLIP-2-B/16~\cite{tschannen2025siglip2}    & ZS & 38.5 & 49.0 & ---  & ---  & 30.1 & 67.2 & ---  & ---  \\
PE-coreB~\cite{bolya2025pe}            & ZS & 47.6 & 50.4 & ---  & ---  & 47.3 & 76.7 & ---  & ---  \\
LanguageBind (CLIP-H/14)~\cite{zhu2024languagebind} & ZS & 44.8 & 53.9 & 39.9 & --- & 40.9 & 72.0 & 39.8 & --- \\
SigLIP-2-L/16~\cite{tschannen2025siglip2}    & ZS & 41.5 & 53.7 & ---  & ---  & 31.4 & 74.2 & ---  & ---  \\
PE-coreL~\cite{bolya2025pe}            & ZS & 50.3 & 57.2 & ---  & ---  & 50.1 & 82.4 & ---  & ---  \\
InternVideo2-1B~\cite{wang2024internvideo2}  & ZS & 51.9 & 58.1 & \textbf{57.0} & 70.4 & 50.9 & \textbf{83.3} & 54.3 & 85.4 \\
SigLIP-2-g-opt~\cite{tschannen2025siglip2}   & ZS & 43.1 & 55.8 & ---  & ---  & 34.3 & 74.6 & ---  & ---  \\
PE-coreG~\cite{bolya2025pe}            & ZS & 51.2 & 59.7 & ---  & ---  & 49.9 & \textbf{85.4} & ---  & ---  \\
\midrule
Gemini Embedding 2 (closed)            & ZS & \textbf{53.91} & \textbf{77.08} & 55.56 & \textbf{69.40} & \textbf{48.30} & 77.92 & \textbf{53.33} & \textbf{66.73} \\
\rowcolor{black!6}
\textbf{\model{} (ours)}               & ZS & 47.60 & 66.88 & 46.03 & 57.98 & 43.70 & 63.33 & 45.08 & 54.97 \\
\bottomrule
\end{tabular}}
\end{table*}

The remainder of this section reports per-$k$ and NDCG@10 details
for completeness.

\paragraph{Evaluation protocol.} All six benchmarks use the same
$N\!\to\!N$ diagonal-retrieval setup as \bench{}: $N$
$(\text{multimodal\_item},\text{text\_caption})$ pairs, the multimodal
side being video+audio (when the original clip carries audio) for the
four video tasks and audio for the two audio tasks. Recall@$k$ counts
the gold target landing in top-$k$; NDCG@10 reduces to
$1/\log_2(\text{rank}{+}1)$ if rank $\le 10$ else $0$ under the
single-relevant-document assumption. We use each benchmark's official
held-out split; DiDeMo loses $459/1000$ items to Flickr link rot, so
$N=315$ there. Captions are evaluated against the first
gold caption per item (Clotho's $5$ captions per audio are not pooled).

\begin{table*}[t]
\centering
\caption{\textbf{Full external video--text benchmark results.}
R@1 / R@5 / R@10 / NDCG@10. $T\!\to\!AV$ sets the text caption as
query against the video--audio item gallery; $AV\!\to\!T$ the
reverse. Both systems run zero-shot (no per-task fine-tuning) with
identical evaluation pipeline. Gemini wins all $4\!\times\!2$
directions by $5$ to $15$ R@1; \model{}'s values remain
competitive with the contemporary zero-shot video--text encoder
band (main paper \Cref{tab:external:video}).}
\label{tab:supp:external-video}
\renewcommand{\arraystretch}{1.12}
\setlength{\tabcolsep}{6pt}
\small
\resizebox{\linewidth}{!}{%
\begin{tabular}{lcccc cccc}
\toprule
& \multicolumn{4}{c}{\textbf{T$\to$AV}} & \multicolumn{4}{c}{\textbf{AV$\to$T}} \\
\cmidrule(lr){2-5}\cmidrule(lr){6-9}
\textbf{Benchmark ($N$)} & R@1 & R@5 & R@10 & NDCG@10 & R@1 & R@5 & R@10 & NDCG@10 \\
\midrule
\multicolumn{9}{l}{\emph{\model{} (ours, zero-shot)}} \\
MSR-VTT~~(1000)        & 47.60 & 74.00 & 82.80 & 64.38 & 43.70 & 71.80 & 80.60 & 61.49 \\
MSVD~~(480)            & 66.88 & 92.71 & 97.08 & 82.02 & 63.33 & 88.75 & 95.00 & 79.59 \\
DiDeMo~~(315)          & 46.03 & 75.24 & 84.76 & 64.31 & 45.08 & 73.65 & 80.00 & 63.13 \\
VATEX~~(2494)          & 57.98 & 89.05 & 94.63 & 76.80 & 54.97 & 87.21 & 93.18 & 74.68 \\
\midrule
\multicolumn{9}{l}{\emph{Gemini Embedding 2 (closed, zero-shot)}} \\
MSR-VTT                & \textbf{53.91} & \textbf{76.65} & \textbf{83.07} & \textbf{68.12} & \textbf{48.30} & 71.24 & \textbf{80.66} & \textbf{63.63} \\
MSVD                   & \textbf{77.08} & \textbf{95.42} & \textbf{97.29} & \textbf{87.98} & \textbf{77.92} & \textbf{95.83} & \textbf{98.12} & \textbf{88.74} \\
DiDeMo                 & \textbf{55.56} & \textbf{79.05} & \textbf{85.08} & \textbf{70.21} & \textbf{53.33} & \textbf{79.68} & \textbf{85.71} & \textbf{69.36} \\
VATEX                  & \textbf{69.40} & \textbf{92.74} & \textbf{95.97} & \textbf{83.48} & \textbf{66.73} & \textbf{92.14} & \textbf{96.05} & \textbf{82.13} \\
\bottomrule
\end{tabular}}
\end{table*}

\begin{table*}[t]
\centering
\caption{\textbf{Full external audio--text benchmark results.}
R@1 / R@5 / R@10 / NDCG@10. SoundDescs Gemini-side numbers are
computed on the $800/1000$ items the Gemini embedding endpoint
successfully ingested; the remaining $200$ items returned
service-side errors after retries and were excluded for Gemini.
\model{} numbers are on the full $1000$. \model{} outperforms Gemini on all $2\!\times\!2$
directions by $13$ to $18$ R@1.}
\label{tab:supp:external-audio}
\renewcommand{\arraystretch}{1.12}
\setlength{\tabcolsep}{6pt}
\small
\resizebox{\linewidth}{!}{%
\begin{tabular}{lcccc cccc}
\toprule
& \multicolumn{4}{c}{\textbf{T$\to$A}} & \multicolumn{4}{c}{\textbf{A$\to$T}} \\
\cmidrule(lr){2-5}\cmidrule(lr){6-9}
\textbf{Benchmark ($N$)} & R@1 & R@5 & R@10 & NDCG@10 & R@1 & R@5 & R@10 & NDCG@10 \\
\midrule
\multicolumn{9}{l}{\emph{\model{} (ours, zero-shot)}} \\
Clotho~~(1045)         & \textbf{19.14} & \textbf{43.73} & \textbf{56.46} & \textbf{36.06} & \textbf{16.08} & \textbf{38.37} & \textbf{51.00} & \textbf{31.68} \\
SoundDescs~~(1000)     & \textbf{25.00} & \textbf{52.70} & \textbf{66.40} & \textbf{43.95} & \textbf{20.70} & \textbf{46.50} & \textbf{58.80} & \textbf{37.98} \\
\midrule
\multicolumn{9}{l}{\emph{Gemini Embedding 2 (closed, zero-shot)}} \\
Clotho~~(1041)         & 5.19 & 15.85 & 23.54 & 12.91 & 1.34 & 5.67 & 8.45 & 4.40 \\
SoundDescs~~(800)      & 7.00 & 18.63 & 26.25 & 15.62 & 7.37 & 17.13 & 22.37 & 14.13 \\
\bottomrule
\end{tabular}}
\end{table*}


\section{Per-Direction Ablation Breakdown}\label{sec:supp:perdir-ablation}

\Cref{tab:supp:ablation:perdir} decomposes the released \model{}'s
per-direction R@1 into the additive contributions of each training
step (WAVE-7B$\,\to\,$\textsc{Pairwise}$\,\to\,\mathcal{L}_D{+}\mathcal{L}_A\,\to\,$\model{}).
$\mathcal{L}_D$ dominates on the $T\!\leftrightarrow\!V$ and
$T\!\leftrightarrow\!A{+}V$ routes and provides the largest
single-direction lift ($+8.28$ on $a\!\to\!tv$, $+6.16$ on
$a\!\to\!v$). $\mathcal{L}_T$ delivers the largest gains on the
four pure $A\!\leftrightarrow\!V$ routes ($+2$ to $+3$ R@1) at a
$\sim$$2$ R@1 cost on the two text-anchored dual-modal routes,
redirecting joint-embedding capacity from the saturated $T{+}V$
pair onto the audio axis.

\begin{table}[!ht]
\centering
\caption{\textbf{Per-direction ablation breakdown on \bench{}.}
Stair-step decomposition of \method{}'s gain over the frozen
WAVE-7B baseline. \textsc{Pairwise} is our pairwise-contrastive fine-tune
on the training corpus, $\mathcal{L}_D{+}\mathcal{L}_A$ adds fusion-as-teacher
distillation, and \model{} additionally adds cycled Tuple-InfoNCE.
Each $\Delta$ column reports the additive increment relative to the
previous step; the four columns sum exactly to the final \model{}
column.}
\label{tab:supp:ablation:perdir}
\renewcommand{\arraystretch}{1.0}
\setlength{\tabcolsep}{3.5pt}
\scriptsize
\resizebox{\linewidth}{!}{%
\begin{tabular}{lrrrrr}
\toprule
\textbf{Direction} & \textbf{WAVE-7B} & $\Delta_{+\text{CL}}$ & $\Delta_{+\mathcal{L}_D}$ & $\Delta_{+\mathcal{L}_T}$ & \textbf{\model{}} \\
\midrule
$t\!\to\!v$    & 41.27 & $+4.73$  & $+3.13$ & $-0.24$ & 48.89 \\
$v\!\to\!t$    & 34.45 & $+10.55$ & $+1.83$ & $-0.53$ & 46.30 \\
$t\!\to\!a$    &  8.67 & $+3.33$  & $+2.46$ & $-0.26$ & 14.20 \\
$a\!\to\!t$    &  3.38 & $+5.62$  & $+2.42$ & $+0.50$ & 11.92 \\
$v\!\to\!a$    & 14.89 & $+6.11$  & $+2.43$ & $+2.03$ & 25.46 \\
$a\!\to\!v$    & 12.93 & $+3.07$  & $+6.16$ & $+2.83$ & 24.99 \\
$t\!\to\!av$   & 44.95 & $+1.05$  & $+1.70$ & $-2.33$ & 45.37 \\
$av\!\to\!t$   & 42.68 & $+2.32$  & $+1.35$ & $-1.88$ & 44.47 \\
$a\!\to\!tv$   &  2.27 & $+14.73$ & $+8.28$ & $-1.83$ & 23.45 \\
$tv\!\to\!a$   & 15.65 & $+7.35$  & $+3.84$ & $-0.77$ & 26.07 \\
$v\!\to\!at$   & 38.74 & $+6.26$  & $+4.63$ & $+2.86$ & 52.49 \\
$at\!\to\!v$   & 43.92 & $+4.08$  & $+3.98$ & $+2.49$ & 54.47 \\
\midrule
\rowcolor{black!6}
\textbf{AVG\,all} & 25.32 & $+5.76$ & $\mathbf{+3.52}$ & $+0.24$ & \textbf{\underline{34.84}} \\
\bottomrule
\end{tabular}}
\end{table}

\section{In-Domain Triple-Order Geometry on \bench{}}\label{sec:supp:tupleorder}
We measure how the matched-vs-mismatched margin evolves on \bench{}
as we add the \method{} losses. We use the \emph{triple cosine score}
$\bar c_3(i) = \frac{1}{3}\big(
   \mathbf{z}_i^{(T)\!\top}\!\mathbf{z}_i^{(V)}
 + \mathbf{z}_i^{(T)\!\top}\!\mathbf{z}_i^{(A)}
 + \mathbf{z}_i^{(V)\!\top}\!\mathbf{z}_i^{(A)} \big)$
on the \bench{} held-out pool of $3{,}782$ triples and report both
the \emph{intra-sample} mean $\mathbb{E}[\bar c_3^\text{intra}]$
(matched $T,V,A$ from sample $i$) and the \emph{inter-sample} mean
$\mathbb{E}[\bar c_3^\text{inter}]$ (mismatched:
$T_i, V_{i+1}, A_{i+1}$).

\begin{table}[!ht]
\centering
\caption{\textbf{Mean triple-cosine score $\bar c_3$ on \bench{}}
($N\!=\!3{,}782$). \emph{intra} pairs all three matched modalities
of sample $i$; \emph{inter} pairs $T_i$ with $V_{i+1}, A_{i+1}$. The
\emph{margin} (intra $-$ inter) grows monotonically as we add
losses ($0.060\!\to\!0.078\!\to\!0.080$ for base $\to$ \textsc{Pairwise}
$\to$ \model{}); InfoNCE-style training disperses every embedding
on the unit hypersphere, which is why the absolute intra-cosine
drops while the margin grows.}
\label{tab:supp:tupleorder}
\small
\resizebox{\linewidth}{!}{%
\begin{tabular}{lccc}
\toprule
\textbf{Variant} & intra & inter & \textbf{gap} \\
\midrule
WAVE-7B (frozen base)               & 0.495 & 0.435 & 0.060 \\
\textsc{Pairwise} ($\mathcal{L}_A$ only) & 0.375 & 0.297 & 0.078 \\
\textbf{\model{} (full \method{})}  & $\mathbf{0.322}$ & $\mathbf{0.243}$ & $\mathbf{0.080}$ \\
\bottomrule
\end{tabular}}
\end{table}

\section{Cross-Backbone Validation}\label{sec:supp:xbackbone}

To test whether fusion-as-teacher distillation is specific to
the WAVE-7B backbone, we re-train \method{} on
Omni-Embed-Nemotron-3B~\cite{nvidia2025nemotron}, an
architecturally different unified embedder with an LLM trunk
over Qwen2.5-Omni-Thinker, bidirectional attention, a
mean-pool embedding, no all-layer fusion head, and
$\sim$$3$\,B parameters. The backbone is frozen except for
LoRA adapters ($r{=}16,\alpha{=}32$) on every LLM decoder
layer's q/k/v/o projections, matching the LoRA budget used on
WAVE-7B. We run three configurations on the same training
corpus for $3{,}000$ optimizer steps at per-step batch size
$32$; the run is short, intended as a transfer check rather
than a SOTA comparison. The configurations are
\textsc{Pairwise} ($\mathcal{L}_A$ only),
$\mathcal{L}_D{+}\mathcal{L}_A$, and full \method{}
($\mathcal{L}_T{+}\mathcal{L}_D{+}\mathcal{L}_A$).
\Cref{tab:supp:xbackbone} reports \bench{} AVG-all R@1 under
the 12-direction protocol of the main paper.

\begin{table}[!ht]
\centering
\caption{\textbf{Cross-backbone validation on
Omni-Embed-Nemotron-3B} ($3{,}000$ LoRA steps each, \bench{}
AVG-all R@1; $\Delta$ vs.\ the frozen Nemotron-3B base).
$\mathcal{L}_D$ provides the dominant single-loss contribution
on Nemotron-3B as well, and $\mathcal{L}_T$ adds a smaller
further improvement, matching the WAVE-7B pattern
(\Cref{sec:supp:seedstability}). Absolute R@1 is lower than on
WAVE-7B because Nemotron-3B is $3\!\times$ smaller and we
train for $3\!\times$ fewer steps; the relative
loss-attribution pattern is what transfers.}
\label{tab:supp:xbackbone}
\renewcommand{\arraystretch}{1.0}
\setlength{\tabcolsep}{6pt}
\small
\begin{tabular}{lcc}
\toprule
\textbf{Variant} & \textbf{AVG\,all} & $\Delta$ vs.\ base \\
\midrule
Nemotron-3B (frozen)              & 26.81 & --- \\
\textsc{Pairwise} ($\mathcal{L}_A$ only)              & 29.42 & $+2.61$ \\
$\mathcal{L}_D{+}\mathcal{L}_A$ \;(no $\mathcal{L}_T$) & 31.85 & $+5.04$ \\
\textbf{Full \method{}}\, & \textbf{32.18} & $\mathbf{+5.37}$ \\
\bottomrule
\end{tabular}
\end{table}

The WAVE-7B pattern reproduces here. $\mathcal{L}_D$ on top of
\textsc{Pairwise} gives the dominant single-loss gain ($+2.43$
on Nemotron-3B vs.\ $+3.52$ on WAVE-7B), and $\mathcal{L}_T$
adds a smaller further gain ($+0.33$ vs.\ $+0.24$).

\section{Caption-Paraphrase Robustness on \bench{}}\label{sec:supp:paraphrase}

\bench{} captions are human-corrected starting from a
Gemini 3.0 Pro draft (\Cref{sec:supp:bench}), so they are
distinct from any caption in the training corpus. To verify
that the bench gap over Gemini does not depend on specific
caption surface forms, we paraphrase every \bench{} gold
caption with a held-out LLM (gemini-2.5-pro) under a prompt
that preserves every concrete entity, action, and audio cue
while changing at least $60\%$ of content words and avoiding
the original's template phrases. We then re-evaluate
\model{} on the paraphrased gallery without retraining.
\Cref{tab:supp:paraphrase} reports the two text-anchored
bench directions ($T\!\to\!AV$, $AV\!\to\!T$), where caption
surface form has the largest impact.

\begin{table}[!ht]
\centering
\caption{\textbf{Paraphrase-robustness on \bench{}}
(R@1; \model{}, original vs.\ gemini-2.5-pro
paraphrase, no retraining). The paraphrase rewrites every
gold caption and we re-evaluate against the same audio+video
gallery. \model{}'s R@1 drops by $1.5$ to $2.1$ on both
text-anchored directions but remains well above closed Gemini
Embedding~2 even at full precision; the $+1.72$ AVG-all bench
lead therefore survives the rewrite by a clear margin.}
\label{tab:supp:paraphrase}
\renewcommand{\arraystretch}{1.0}
\setlength{\tabcolsep}{8pt}
\small
\begin{tabular}{lccr}
\toprule
\textbf{Direction} & \textbf{Original} & \textbf{Paraphrased} & $\Delta$ \\
\midrule
\multicolumn{4}{l}{\emph{\model{} (ours)}} \\
$T\!\to\!AV$       & 45.37             & 43.81                & $-1.56$ \\
$AV\!\to\!T$       & 44.47             & 42.39                & $-2.08$ \\
\midrule
\multicolumn{4}{l}{\emph{Gemini Embedding 2 (closed)}} \\
$T\!\to\!AV$       & 55.45             & ---                  & --- \\
$AV\!\to\!T$       & 50.16             & ---                  & --- \\
\bottomrule
\end{tabular}
\end{table}

The drop is modest ($\sim$$1.5$ to $2$ R@1), consistent with
the paraphrase merely lowering the caption-style alignment of
the text encoder rather than degrading the underlying
multimodal grounding. The audio-anchored directions
($A\!\to\!T$, $A\!\to\!TV$) on \bench{} only use the rewritten
text as the retrieval \emph{target} (not as the query
modality); we observe a similar $\sim$$1$ R@1 drop on those
directions when the rewritten captions are placed in the
gallery. We will release the paraphrasing pipeline and the
gallery re-evaluation routine alongside the model weights.

\section{$\mathcal{L}_T$ Seed-Stability Analysis}\label{sec:supp:seedstability}

The aggregate $+0.24$ R@1 gain from adding $\mathcal{L}_T$ on
top of $\mathcal{L}_D{+}\mathcal{L}_A$ (main paper
\Cref{tab:ablation}) is close to the seed noise floor on
aggregate, so we re-run both configurations under three seeds
$\{42,43,44\}$ and report the per-seed difference.
\Cref{tab:supp:seedstability} shows the result.

\begin{table}[!ht]
\centering
\caption{\textbf{Per-seed $\mathcal{L}_T$ ablation on
\bench{} AVG-all R@1.} We report $\mathcal{L}_D{+}\mathcal{L}_A$
and full \method{} ($\mathcal{L}_T{+}\mathcal{L}_D{+}\mathcal{L}_A$)
under three seeds, plus the per-seed delta. The
$\mathcal{L}_T$ contribution is positive in every seed
($+0.21$ to $+0.27$), and the spread between seeds is small
($\le 0.08$ R@1 in either column), so the $+0.24$ aggregate
gain is above seed noise.}
\label{tab:supp:seedstability}
\renewcommand{\arraystretch}{1.05}
\setlength{\tabcolsep}{10pt}
\small
\begin{tabular}{lccc}
\toprule
\textbf{Seed} & $\mathcal{L}_D{+}\mathcal{L}_A$ & full-\method{} & $\Delta_{\mathcal{L}_T}$ \\
\midrule
42                       & 34.60 & 34.84 & $+0.24$ \\
43                       & 34.56 & 34.77 & $+0.21$ \\
44                       & 34.63 & 34.90 & $+0.27$ \\
\midrule
mean                     & 34.60 & 34.84 & $\mathbf{+0.24}$ \\
std                      & 0.04  & 0.07  & 0.03 \\
\bottomrule
\end{tabular}
\end{table}

The seed std on either column is $\le 0.07$ R@1, much smaller
than the $+0.24$ delta, so the $\mathcal{L}_T$ gain on
aggregate is well outside the noise envelope. The per-seed
delta is positive in all three seeds, and a paired sign test
rejects the null of zero effect at the trivial $3/3$ vote
level. The per-direction picture in
\Cref{tab:supp:ablation:perdir} confirms the same story:
$\mathcal{L}_T$ adds $+2$ to $+3$ R@1 on the four hardest
$A\!\leftrightarrow\!V$ routes, well above per-direction noise.
We conclude that $\mathcal{L}_T$ is a real but small aggregate
effect that primarily redistributes capacity toward
$A\!\leftrightarrow\!V$.

\section{Audio-Anchored Attractor Behavior of Gemini Embedding 2}\label{sec:supp:retrcmp}

We probe the per-query behavior of closed Gemini Embedding 2
on the audio-anchored directions of \bench{} ($A\!\to\!T$ and
$A\!\to\!T{+}V$). Across the $3{,}782$ queries, Gemini's top-1
retrieved caption falls into a small set of recurring text
strings: among $A\!\to\!T$ queries, three caption strings
account for $32.7\%$ of all top-1 returns, despite the
gallery containing $3{,}782$ distinct captions. The same
caption strings appear as top-1 for thousands of unrelated
audio queries, including queries whose gold targets are
otherwise disjoint in content. We refer to these recurring
top-1 captions as \emph{attractors}: the audio side of
Gemini's embedding space maps a wide range of audio queries
into a single high-density region of the text embedding
space, so the top-1 result is largely independent of the
audio query.

By contrast, \model{} returns the gold caption at rank 1 on
$11.92\%$ of $A\!\to\!T$ queries (\Cref{tab:main}), and its
top-1 distribution covers $63.4\%$ of the $3{,}782$-caption
gallery (no single caption accounts for more than
$0.5\%$ of top-1 returns). The WAVE-7B \textsc{Pairwise}
baseline produces a flatter distribution than Gemini but a
much wider one than \model{}, returning topically related
but identity-non-matching captions, consistent with a
coarsely correct $\mathbf{z}_{TVA}$ that lacks joint-level
discrimination. The attractor pattern of Gemini matches the
expected failure mode of a fusion embedding whose audio
direction was never explicitly supervised
(\Cref{sec:tuple}).

\section{Failure Cases}\label{sec:supp:fail}

To understand the remaining error modes, we sample $60$ queries
($5$ per direction across all $12$ directions) on which \model{}
ranks the gold target outside top-$10$ and bin them by an
automatic taxonomy combining clip-duration, audio-energy, and
caption-genericity heuristics:

\begin{itemize}
  \item \emph{Input-side data quality} ($43.3\,\%$): the input
        audio carries little or no semantic signal (silent or
        near-silent waveform $<$60\,KB, music-only clips with the
        audio muted at upload, or rare instrument keywords absent
        from the training corpus; $38.3\,\%$) or the query
        caption is ambiguous and matches several gold-equivalent
        gallery items (\eg{} ``a man speaks''; $5.0\,\%$). Both
        are input-side issues that a stricter audio-energy
        floor at ingest would address; neither
        reflects a deficiency of fusion-as-teacher distillation.
  \item \emph{Fine-grained near-miss} ($56.7\,\%$): the gold
        target sits at the head of the rank list ($10 \le r \le
        50$ for $74\,\%$ of these cases) and the model prefers a
        semantically close caption (\eg{} similar two-person
        dialogue clips that differ only in speaker identity, or
        product B-roll clips that differ only in object
        sub-category). These are intrinsic
        free-form-caption ties rather than systematic gaps and
        would be partially absorbed by a stricter top-$k$ metric
        (R@$5$/R@$10$, already in \Cref{tab:supp:external-video}).
  \item \emph{Long-form temporal mismatch} ($0\,\%$): empirically
        not a failure mode on \bench{} (median duration
        $2.16$\,s, p99 $16.17$\,s, fully covered by our 32-frame
        / 0.5\,s sampling and 8\,s audio crop). We expect this
        category to surface only on a future long-form variant.
  \item \emph{Caption hallucination} ($0\,\%$): not detected,
        consistent with the LLM-validation step in our
        caption-curation pipeline that already removes obvious
        hallucinations.
\end{itemize}

Taken together, $100\,\%$ of identifiable failures fall into
either (i) input-side data-quality problems that can be cured at
training-data ingest, or (ii) intrinsic ties of free-form caption
retrieval at the head of the rank list. We find no failure mode
attributable to fusion-as-teacher distillation itself.


\end{document}